\documentclass[letterpaper,twocolumn,10pt]{article}
\usepackage{usenix-2020-09}

\usepackage{tikz}
\usepackage{amsmath}

\usepackage{xurl}
\usepackage{xspace}
\usepackage{subcaption}
\usepackage{graphicx}
\usepackage{xcolor}
\usepackage{booktabs}
\usepackage{pifont}
\usepackage{comment}
\usepackage{enumitem}
\usepackage{makecell} 

\usepackage[normalem]{ulem}

\usepackage[ruled,linesnumbered, noend]{algorithm2e}
\SetKwComment{Comment}{/* }{ */}

\newcommand{\sys}{ServerlessLLM\xspace}

\ifdefined\RELEASE
  \newcommand{\ignore}[1]{}
  \newcommand{\fixme}[1]{}
  \newcommand{\yao}[1]{}
  \newcommand{\andrei}[1]{}
  \newcommand{\luo}[1]{}
  \newcommand{\dmitrii}[1]{}
  \newcommand{\leyang}[1]{}
  \newcommand{\TODO}[1]{}

\else
  \newcommand{\ignore}[1]{}
  \newcommand{\fixme}[1]{{\textcolor{red}{[~FIXME:~#1~]}}}
  \newcommand{\yao}[1]{{\textcolor{orange}{[~YAO:~#1~]}}}
  \newcommand{\andrei}[1]{{\textcolor{olive}{[~AND:~#1~]}}}
  \newcommand{\luo}[1]{{\textcolor{gray}{[~LUO:~#1~]}}}
  \newcommand{\dmitrii}[1]{{\textcolor{blue}{[~DMI:~#1~]}}}
  \newcommand{\leyang}[1]{{\textcolor{teal}{[~LEY:~#1~]}}}
  \newcommand{\TODO}[1]{{\textcolor{red}{TODO:~#1}}}

\fi

\usepackage{glossaries}
\makeglossaries
\usepackage{tikz}

\newcommand{\tinyskip}{\vspace{3pt}}
\newcommand{\mypar}[1]{\tinyskip\noindent\textbf{#1.}\xspace}
\newcommand{\myitem}[1]{\item\textbf{#1.}\xspace}

\newcommand*\myc[1]{%
\scalebox{0.78}{\begin{tikzpicture}[baseline=-3pt]
  \node[draw,circle,inner sep=0.5pt, fill=black] {\textcolor{white}{\textsf{\textbf{#1}}}};
\end{tikzpicture}}}

\usepackage{mathtools}
\newcommand{\eg}{\text{e.g.,}\ }

\newcommand{\ie}{\text{i.e.,}\ }

\newcommand{\todo}[1]{\textbf{\color{red}[#1]}}

\usepackage{xcolor}

\begin{document}

\date{}

\title{\fontsize{14}{15}\selectfont \sys: Low-Latency Serverless Inference for Large Language Models}

\author{
{\rm Yao Fu$^1$ Leyang Xue$^1$ Yeqi Huang$^1$ Andrei-Octavian Brabete$^1$ Dmitrii Ustiugov$^2$ Yuvraj Patel$^1$ Luo Mai$^1$ } \\ \vspace{-.5em} \\ $^1$University of Edinburgh \quad $^2$NTU Singapore
}
\maketitle

\begin{abstract}


This paper presents \sys, a distributed system designed to support low-latency serverless inference for Large Language Models (LLMs). By harnessing the substantial near-GPU storage and memory capacities of inference servers, \sys 
achieves effective local checkpoint storage, minimizing the need for remote checkpoint downloads and ensuring efficient checkpoint loading. The design of \sys features three core contributions:
(i) \emph{fast multi-tier checkpoint loading}, featuring a new loading-optimized checkpoint format and a multi-tier loading system, fully utilizing the bandwidth of complex storage hierarchies on GPU servers; (ii) \emph{efficient live migration of LLM inference}, which enables newly initiated inferences to capitalize on local checkpoint storage while ensuring minimal user interruption; and (iii) \emph{startup-time-optimized model scheduling}, which assesses the locality statuses of checkpoints on each server and schedules the model onto servers that minimize the time to start the inference. Comprehensive evaluations, including microbenchmarks and real-world scenarios, demonstrate that \sys dramatically outperforms state-of-the-art serverless systems, reducing latency by 10 - 200X across various LLM inference workloads.

\end{abstract}

\section{Introduction}

Large Language Models (LLMs) have recently been integrated into various online applications, such as programming assistants~\cite{copilot}, search engines~\cite{new-bing}, and conversational bots~\cite{chatgpt}. These applications process user inputs, such as questions, by breaking them down into tokens (e.g., words). LLMs generate responses in an autoregressive manner, predicting each subsequent token based on the combination of input tokens and those already generated, until a sentence-ending token (EoS) is reached. To optimize this process, LLMs utilize key-value caches to store intermediate results, thereby minimizing redundant computations.

Serving LLMs at scale presents significant challenges due to the extensive GPU resources required and the stringent low response time constraints demanded by interactive services. Additionally, LLM inference latency is unpredictable as it depends on the output length, which varies significantly due to iterative token generation~\cite{wu2023fast, vllm, databricks-blog}.

To achieve low latency, processing an LLM request often requires multiple GPUs for durations ranging from seconds to minutes. In practice, service providers hosting LLMs need to cater to a diverse range of developers, leading to substantial GPU consumption~\cite{bananadevtechnicalblog} and impacting the sustainability of LLM services~\cite{DBLP:conf/hotcarbon/ChienLNRSW23}. Consequently, LLM inference services are compelled to impose strict limits on the number of requests users can send (e.g., 40 messages per 3 hours for ChatGPT~\cite{chatgpt}), highlighting the providers' current challenges in meeting demand. Researchers predict that LLM inference costs could escalate by more than 50 times as it approaches the popularity of Google Search~\cite{DBLP:conf/hotcarbon/ChienLNRSW23}.

To reduce GPU consumption, LLM service operators are turning to serverless inference, as demonstrated in platforms such as Amazon SageMaker~\cite{sagemaker}, Azure~\cite{azureml}, KServe~\cite{kserve}, and HuggingFace~\cite{huggingface}. 
In this model, developers upload their LLM checkpoints, which include both model execution and parameter files, to a checkpoint storage system. When a request is received, a model loading scheduler selects available GPUs to initiate these checkpoints. A request router then directs the inference request to the selected GPUs. This serverless approach allows infrastructure providers to efficiently multiplex LLMs on GPUs, improving resource utilization. Additionally, it offers economic advantages to infrastructure users, who incur costs only for each request's duration, thereby avoiding expensive long-term GPU reservations.


While serverless inference offers cost savings for deploying LLMs, it also introduces significant latency overheads. These overheads are commonly attributed to inference cold starts, a frequent issue in serverless workloads, as demonstrated by public traces~\cite{254430, 10.1145/3477132.3483580}. Cold starts are especially prolonged for LLM checkpoints, whose sizes can range from gigabytes~\cite{touvron2023llama2, falcon40b, DBLP:journals/corr/abs-2205-01068} to terabytes~\cite{switchtransformer}. The vast size is due to the immense number of parameters in such models, leading to notable delays when downloading from remote storage. Moreover, these checkpoints consist of numerous tensors, each with unique structures and sizes. The complex process of loading these tensors onto GPUs, which involves file deserialization, memory allocation, and tensor shape parsing, further compounds these delays.

We aim to explore system designs that support low-latency serverless inference for LLMs. We note that GPU-based inference servers typically feature a sophisticated yet underutilized storage hierarchy, equipped with extensive host memory and storage capacities. Current serverless inference systems, such as KServe~\cite{kserve} and Ray Serve~\cite{ray-serve}, often only utilize a fraction of the available host memory and minimally employ SSDs for caching checkpoints from the model repository. This observation has led us to propose a novel system design: leveraging the multi-tier storage hierarchy for local checkpoint storage and harnessing their significant storage bandwidth for efficient checkpoint loading.

However, several open concerns arise when implementing local checkpoint storage:
(i) Given the complex storage architecture of a GPU server, which includes multiple GPUs, DRAM, SSDs, and remote storage, all interconnected through various links such as PCIe, NVMe, and network connections, how can we optimize the loading of LLM checkpoints to fully exploit the available bandwidth? 
(ii) Assigning requests to servers with pre-loaded checkpoints can avoid the need for remote checkpoint downloads, but this strategy might lead to prolonged queuing delays or high preemption costs. This is particularly challenging as LLMs typically have long, unpredictable inference durations, which differ markedly from traditional deep neural network inference.
(iii) In a distributed cluster where model requests are concurrently served and checkpoints are preloaded onto various layers of local storage, which servers should be strategically selected to minimize the time required to start a model inference?

To address the above, we have designed and implemented \sys, which includes three core contributions:

\mypar{(1) Fast multi-tier checkpoint loading}
\sys can maximize
the storage bandwidth usage of GPU servers for LLM checkpoint loading. It introduces (i) a new \emph{loading-optimized
checkpoint} that supports sequential, chunk-based reading and
efficient tensor in-memory addressing, and (ii) an \emph{efficient multi-tier checkpoint storage system} that can harness the substantial capacity and bandwidth on a multi-tier storage hierarchy, through an in-memory data chunk pool, memory-copy efficient data path, and a multi-stage data loading pipeline.

\mypar{(2) Efficient live migration of LLM inference}
We motivate the need for live migration of LLM inference and are the first to implement LLM live migration in serverless inference systems
to enhance the performance when supporting locality-driven
inference. To achieve high efficiency when migrating LLM inference, we have implemented two strategic designs: (i) the source server migrates only the tokens, rather than the large kv-cache, which significantly reduces network traffic during the migration; and (ii) it triggers an efficient re-computation of the kv-cache at the destination server, ensuring migration can complete in a timely manner.

\mypar{(3) Startup-time-optimized model scheduling}
\sys aids serverless inference systems by enabling latency-preserving, locality-aware model scheduling. It integrates cost models for accurately estimating the time of loading checkpoints from different tiers in the storage hierarchy and the time of migrating an LLM inference to another server. Based on the estimation results, Phantom can choose
the best server to minimize model startup latency.

We have conducted comprehensive evaluation to compare \sys against various baseline methods in a GPU cluster. Micro-benchmark results revealed that \sys's LLM checkpoint loading significantly outperforms existing systems such as Safetensors~\cite{safetensors}, and PyTorch~\cite{pytorch}, achieving loading times that are 3.6 - 8.2X faster. This performance enhancement is particularly notable with large LLMs like OPT~\cite{DBLP:journals/corr/abs-2205-01068}, LLaMA-2~\cite{touvron2023llama2}, and Falcon~\cite{falcon40b}. \sys also supports emerging LoRA adaptors~\cite{hu2021lora}, achieving 4.4X speed-ups in checkpoint loading.

Furthermore, we evaluated \sys with real-world serverless workloads, modeled on the public Azure Trace~\cite{254430}, and benchmarked it against KServe, Ray Serve, and a Ray Serve variant with local checkpoint caching. In these scenarios, \sys demonstrated a 10 to 200 times improvement in latency for running OPT model inferences across datasets (\ie GSM8K~\cite{DBLP:journals/corr/abs-2110-14168} and ShareGPT~\cite{sharegpt4}). These results underscore \sys's effectiveness in combining fast checkpoint loading, efficient inference migration, and optimized scheduling for model loading. The source code for \sys is released at \url{https://github.com/ServerlessLLM/ServerlessLLM}.
\section{Background and Motivation}





\subsection{Why Serverless Inference for LLMs}

Numerous companies, including Amazon, Azure, Google, HuggingFace, Together AI~\cite{together_ai_serverless}, Deepinfra~\cite{deepinfra}, Replicate~\cite{replicate}, Databricks~\cite{databricks}, Fireworks-AI~\cite{fireworks}, and Cohere~\cite{cohere}, have introduced serverless inference services (also known as serverless model entrypoints). These services enable users to deploy standard open-source LLMs either in their original form or by modifying them through fine-tuning or by running custom-built models.

Serverless inference can significantly reduce costs for LLM users by charging only for the duration of inference and the volume of processed data. These serverless platforms also offer functionalities such as auto-scaling and auto-failure-recovery to keep instances in an "always-on" state. For the providing companies, serverless inference allows effective multiplexing of models within a GPU cluster, improving resource utilization, and generating a software premium for managing infrastructure on behalf of users.

Serverless inference systems are especially advantageous for LLM applications with dynamic and unpredictable workloads. These may include newly launched products without clear predictions of user engagement (e.g., the launch of the ChatGPT service) or those facing spontaneous and unpredictable demands, which are typical in sectors such as healthcare, education, legal, and sales. Unlike global-scale LLM services, these applications are activated only when users access the LLM service.



\begin{figure}[t]
    \centering
    \includegraphics[width=\linewidth]{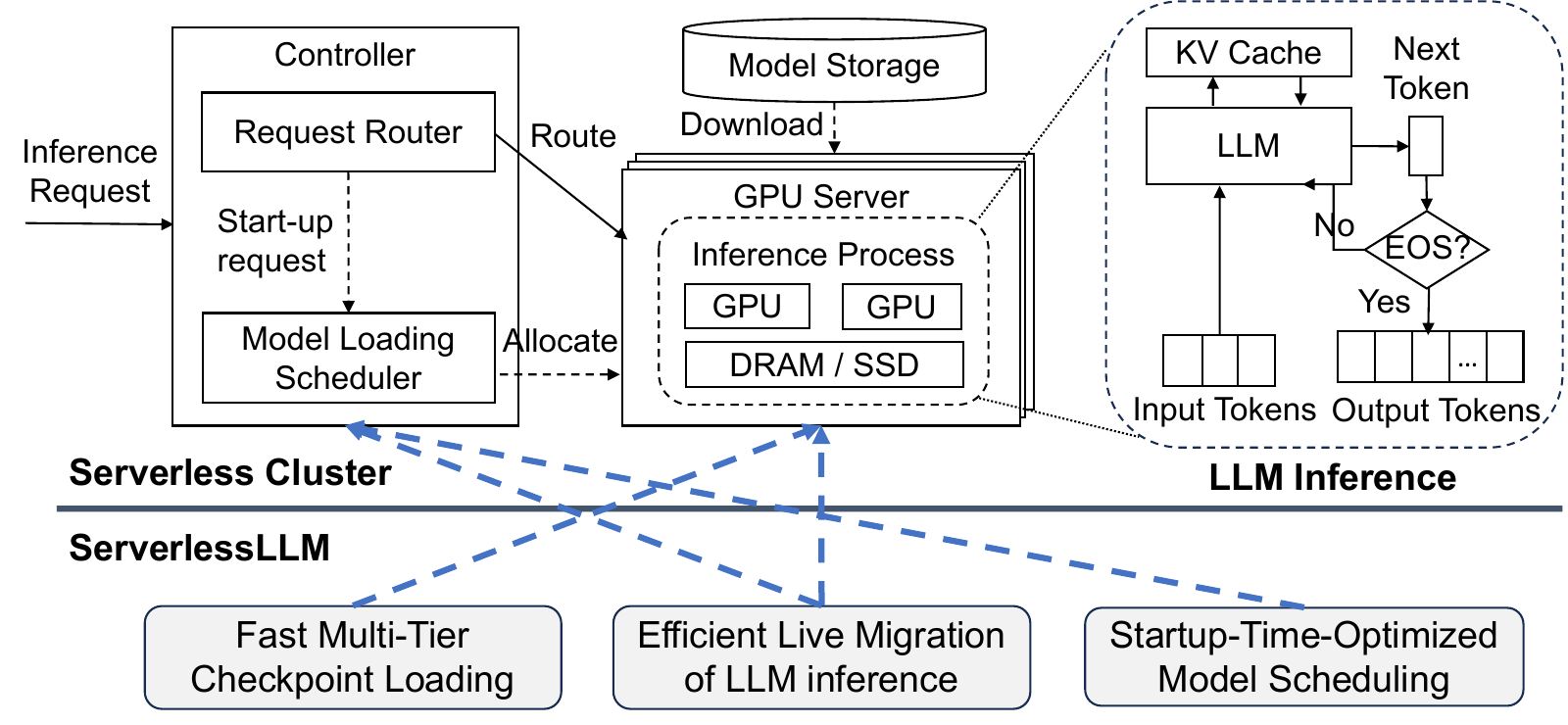}
    \caption{Overview of GPU serverless clusters, LLM inference and new designs introduced by \sys.}
    \label{fig:llm-inference-cluster-overview}
\end{figure}

\subsection{Serverless Cluster and LLM Inference}

We introduce the key components in GPU serverless clusters in Figure~\ref{fig:llm-inference-cluster-overview}. Upon receiving a new inference request, the controller dispatches it to GPU-equipped nodes in a cluster running LLM inference service instances, and to cloud storage hosting model checkpoints. The controller typically consists of two main components: the \emph{request router} and the \emph{model loading scheduler}. The request router directs incoming requests to nodes already running LLM inference processes, or instructs the model loading scheduler to activate LLM inference processes on unallocated GPUs. The selected GPU node initiates a GPU process/container, setting up an inference library (\eg HuggingFace Accelerate~\cite{accelerate} and vLLM~\cite{vllm}). This inference process involves downloading the requested model’s checkpoint from a remote model storage and loading it into the GPU, passing through SSD and DRAM.

The LLM inference process often handle requests that include user-specified input prompts, i.e., a list of tokens, as shown in Figure~\ref{fig:llm-inference-cluster-overview}. This process iteratively generates tokens based on the prompt and all previously generated tokens, continuing until an end-of-sentence token (denoted as EoS) is produced, resulting in non-deterministic total inference time~\cite{patel2023splitwise}. During each iteration, the LLM caches intermediate computations in a KV-cache to accelerate subsequent token generation~\cite{vllm, pope2023efficiently}. The tokens generated by each iteration are continuously streamed back to the requesting client, making LLM applications interactive by nature. Their performance is thus measured by both first-token latency (\ie the time to return the first token) and per-token latency (\ie the average time to generate a token).

\subsection{Challenges with Serverless LLM Inference}
\label{sec:problems}



The deployment of LLMs on serverless systems, although promising, often incurs significant latency overheads. This is largely due to the substantial proportions of cold-start in serverless clusters, as demonstrated by public data: the Azure Trace~\cite{254430} shows that over 40\% of functions exhibit a cold-start rate exceeding 25\%, and approximately 25\% of functions experience a cold-start rate greater than 60\%, within a 5-minute keep-alive interval. These figures align with the findings from our experiments, underscoring the impact of cold-starts in real-world settings. Consequently, many serverless providers, including Bloomberg, have publicly acknowledged experiencing extremely high latencies, often reaching tens of seconds, when initializing state-of-the-art LLMs for inference on their platforms.

We observe several primary reasons for the prolonged LLM cold-start latency:

\mypar{(1) LLM checkpoints are large, prolonging downloads}
LLM checkpoints are significantly larger than conventional DNN checkpoints, which leads to longer download times. For instance, Grok-1~\cite{grok-1} checkpoints are over 600~GB, DBRX~\cite{dbrx-2024} are 250GB, and Mixtral-8x22B~\cite{mistral-8x22b-2024} are about 280GB~\footnote{Model size calculated in float16 precision.} 
Downloading such large checkpoints from remote storage becomes costly. For example, acquiring an LLM checkpoint with a size of 130GB (e.g., LLaMA-2-70B~\cite{touvron2023llama2}) from S3 or blob storage takes a minimum of 26 seconds using a fast commodity network capable of 5GB/s~\cite{anyscaletechnicalblog}.

\mypar{(2) Loading LLM checkpoints incurs a lengthy process}
Even when model checkpoints are stored locally on NVMe SSDs, loading these checkpoints into GPUs remains a complex process including model initialization, GPU memory allocation, tensor creation, and tensor data copy, typically taking tens of seconds (as detailed in \S\ref{sec:evaluation_checkpoint}). For instance, loading the OPT-30B model into 4 GPUs requires 34 seconds using PyTorch, and loading LLaMA-2-70B into 8 GPUs takes 84 seconds. This loading latency far exceeds the time required for generating a token during the inference process, which is usually less than 100ms~\cite{patel2023splitwise}. Consequently, the prolonged first-token latency can significantly disrupt user experience. 

\subsection{Existing Solutions and Associated Issues} 

To improve the latency performance when supporting LLMs, existing solutions show a variety of issues:


\mypar{(1) Over-subscribing GPUs} The prevalent solutions~\cite{aws-serverless-inference-cold-starts, 10.1145/3503222.3507709}, aimed at circumventing model download and loading times in serverless inference clusters, frequently involve over-subscribing GPUs to accommodate peak demand scenarios. For instance, AWS Serverless Inference~\cite{aws-serverless-inference-cold-starts} maintains a certain number of GPU instances in a warmed state to alleviate the impacts of slow cold starts. While this strategy is effective for managing conventional smaller models, such as ResNet and BERT, it proves challenging for LLMs, which require substantially greater resources from costly GPUs.


\mypar{(2) Caching checkpoints in host memory} Several solutions~\cite{eurosys23servinggpudirecthostaccess, clockwork} have been developed that cache model checkpoints in the host memory of GPU servers to eliminate the need for model downloads. This approach is typically effective for smaller conventional models (e.g., up to a few GBs~\cite{eurosys23servinggpudirecthostaccess}). However, solely relying on host-memory-based caching proves inadequate for LLMs. LLMs can easily exceed hundreds of GBs in size, challenging the capacity of host memory to store a sufficient number of their checkpoints adequately. The limited size of host memory leads to significant cache misses, resulting in frequent model downloads, as further discussed in \S\ref{sec:evaluation_end2end}.


\mypar{(3) Deploying additional storage servers} Various strategies~\cite{anyscaletechnicalblog} recommend the deployment of additional storage servers within a local cluster to cache model checkpoints. Despite these enhancements, recent trace studies~\cite{anyscaletechnicalblog} indicate that model downloads can exceed 20 seconds through an optimized pipeline, even when connected to local commodity storage servers equipped with a 100 Gbps NIC. Although the integration of faster networks (e.g., 200 Gbps Ethernet or InfiniBand) could reduce this latency, the associated costs of implementing additional storage servers and high-bandwidth networks are substantial~\cite{bai2023empowering, gao2021cloud}. For instance, utilizing network-optimized AWS ElasticCache servers~\cite{aws_elasticache_pricing} to support a 70B model can lead to a 100\% increase in costs. Specifically, cache.c7gn.16xlarge servers, which provide 210GB of memory and 200 Gbps of network performance, are priced at \$16.3/h, equivalent to the cost of an 8-GPU g5.48xlarge server.

\section{Exploiting In-Server Multi-Tier Storage}

\sys addresses the challenges highlighted in the previous sections—namely, high model download times and lengthy model loading—using a design approach that is cost-effective, scalable, and long-term viable.

\subsection{Design Intuitions}

Our design is inspired by the simple observation that GPU servers used for inference feature a multi-tier storage hierarchy with substantial capacity and bandwidth. From a capacity standpoint, these servers are equipped with extensive memory capabilities. For example, a contemporary 8-GPU server can support up to 4 TBs of main memory, 64 TBs on NVMe SSDs, and 192 TBs on SATA SSDs~\cite{dgx-h100}. Additionally, we observe that in the serverless inference context, a significant portion of the host memory and storage devices in GPU servers remains underutilized. 

Regarding bandwidth, GPU servers typically house multiple GPUs, each connected to the host memory via a dedicated PCIe connection, providing significant aggregated bandwidth between the memory and GPU. NVMe and SATA SSDs also connect through their respective links and can be configured in RAID to enhance throughput. For instance, an 8-GPU server utilizing PCIe 5.0 technology can achieve an aggregated bandwidth of 512 GB/s between the host memory and GPUs, and around 60 GB/s from NVMe SSDs (RAID 0) to host memory. 


Building on these observations, we propose a design approach that leverages the unused in-server multi-tier storage capacity to store models locally and load them more rapidly, thus reducing latency. This approach is (i)~\emph{cost-effective}, as it reutilizes existing, underutilized storage resources in GPU servers; (ii)~\emph{scalable}, given that the available local storage capacities and bandwidth can naturally increase with the addition of more inference servers; and (iii)~\emph{long-term viable}, as upcoming GPU servers will include even greater capacities and bandwidth (\emph{e.g.}, each Grace-Hopper GPU features 1 TB on-chip DRAM and a 900GB/s C2C link between on-chip DRAM and HBM).

\subsection{Design Concerns and Overview}

In implementing our design, we identify three crucial concerns that must be addressed.

\mypar{(1) Support complex multi-tiered storage hierarchy} Current checkpoint and model loading tools such as PyTorch~\cite{pytorch}, TensorFlow~\cite{tensorflowSavedModelLoad}, and ONNX Runtime~\cite{onnxruntimeLoad} are primarily designed to enhance the training and debugging phases of model development. However, these tools are not optimized for read performance, which becomes critically important in a serverless inference environment. In these settings, model checkpoints are stored once but need to be frequently loaded and accessed across multiple GPUs. This insufficient optimization for read operations results in significant loading delays. While solutions like Safetensors~\cite{safetensors} can enhance loading performance, as demonstrated in Section~\ref{evaluation}, they still fail to fully leverage the capabilities of a multi-tiered storage hierarchy.

\mypar{(2) Strong locality-driven inference} Supporting efficient model loading alone is insufficient; we also need approaches that can effectively schedule requests onto GPU servers with locally stored checkpoints. Implementing locality-driven LLM inference, however, presents challenges. Current ML model serving systems such as ClockWork~\cite{clockwork} and Shepherd~\cite{shepherd} take checkpoint locality into account. Yet, they either depend on accurate predictions of model inference time, which is problematic with LLMs, or they preempt ongoing model inferences, causing significant downtime and redundant computations. Therefore, \sys must adopt a new approach that is tailored to the unique characteristics of LLM inference (\emph{i.e.}, this workload is interactive and features long, unpredictable durations), necessitating the support for inference live migration, which is further detailed in Section~\ref{sec:migration}.

\mypar{(3) Scheduling models for optimized startup time} \sys is designed to minimize the model startup latency.
The cluster scheduler (or controller) plays a crucial role in scheduling models onto GPU resources to answer incoming inference requests. However, the scheduler needs to carefully consider the checkpoint's locality in the entire cluster. Many factors may influence the overall startup latency,
such as the difference in the bandwidth offered by each layer
in the memory hierarchy. There may be instances where it
is beneficial to move the current inference execution to a
new GPU than to allocate the request to a GPU where the
model may have to be loaded from the storage media. Hence,
\sys needs to accurately estimate the startup times considering the cluster's checkpoint locality status and accordingly allocate resources
to minimize startup time. 

\textbf{Overview.} \sys addresses these concerns with three novel designs, as depicted in Figure~\ref{fig:llm-inference-cluster-overview}. Firstly, it facilitates fast multi-tier checkpoint loading (Section~\ref{checkpoint-loading}) to fully utilize the storage capacity and bandwidth of each GPU server. It also coordinates GPU servers and the cluster controller for efficient live migration of LLM inference (Section~\ref{sec:migration}), ensuring locality-driven inference with minimal resource overhead and user disruption. Lastly, \sys features a startup-time-optimized model scheduling policy (Section~\ref{sec:allocation}) implemented in its controller, effectively analyzing the checkpoint storage status of each server within a cluster, and it chooses a server for initiating a model, minimizing its startup time.

\section{Fast Multi-Tier Checkpoint Loading} \label{checkpoint-loading}

In this section, we introduce the design of fast multi-tier checkpoint loading in \sys, with several key objectives: (i) to fully utilize the bandwidth and capacity of multi-tier local storage on GPU servers, (ii) to ensure predictable loading performance, critical for \sys's readiness in low-latency inference clusters, and (iii) to maintain a generic design that supports checkpoints from various deep learning frameworks.

\subsection{Loading-Optimized Checkpoints}\label{sec:model checkpoint}








\begin{figure}[t]
    \centering
    \includegraphics[width=\linewidth]{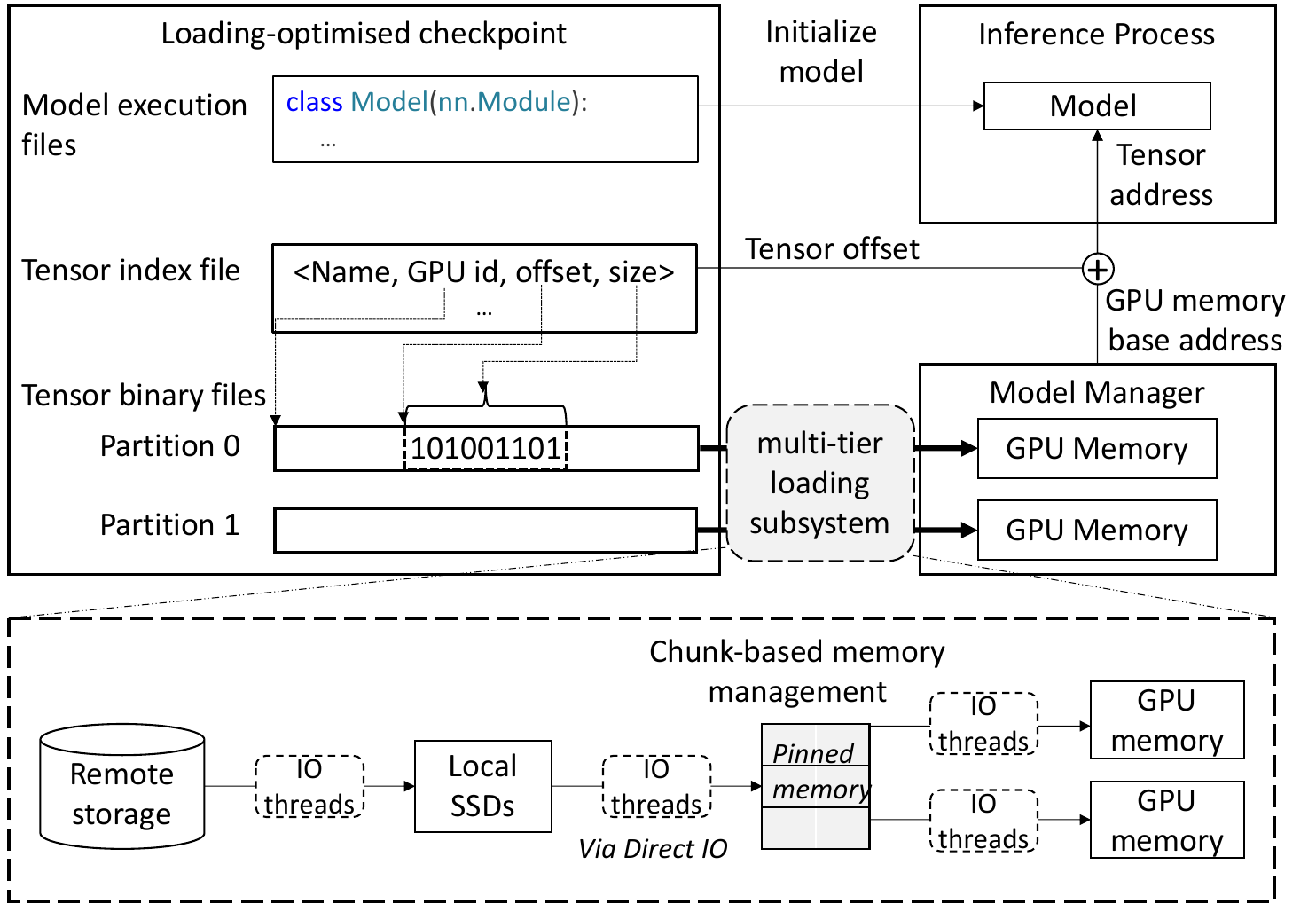}
    \caption{Components in fast multi-tier checkpoint loading.}
    \label{fig:checkpoint loader}
\end{figure}

Our design is motivated by the observation that LLM checkpoints are often written frequently during training and debugging but loaded infrequently. Conversely, in serverless inference environments, checkpoints are uploaded once and loaded multiple times. This discrepancy has inspired us to convert these checkpoints into a loading-optimized format.

To ensure our design is generic for different frameworks, we operate under a set of assumptions that are common in checkpoints. The checkpoints have: (i)~\emph{Model execution files} which define the model architecture.
Depending on the framework, the format varies; TensorFlow typically uses protobuf files \cite{tensorflowmodelformat}, while PyTorch employs Python scripts \cite{torchmodelformat}. Beyond architecture, these files detail the size and shape of each tensor and include a model parallelism plan. This plan specifies the target GPU for each tensor during checkpoint loading. (ii)~\emph{Model parameter files} which stores the binary data of parameters in an LLM.
Tensors within these files can be arranged in any sequence. Runtimes such as PyTorch may also store tensor shapes as indices to calculate the offset and size for each tensor.


To ensure fast loading performance, we implement two main features for the converted checkpoints: (i)~\emph{Sequential chunk-based reading}: To ensure efficient sequential reading, tensors for each GPU are grouped in partitions (shown in Figure~\ref{fig:checkpoint loader}). These files contain only the binary data of model parameters and exclude metadata such as tensor shapes, facilitating large chunk reading. (ii)~\emph{Direct tensor addressing}: We create 
a tensor index file (shown in Figure~\ref{fig:checkpoint loader}) that maps
tensor names to a tuple of GPU id, offset, and size, facilitating the efficient restoration of tensors. The tensors are aligned with memory word sizes, facilitating direct computation of memory address. 

We observe that decoupling the loading and inference processes can further enhance loading performance. This separation allows checkpoint loading to be pre-scheduled and overlapped with the initialization of the inference process. For this, \sys uses a model manager to load tensor data, while allowing the inference process to focus on initializing the model by setting the data pointers for each tensor. More specifically, the model manager allocates memory on GPUs and loads the binary data of the checkpoint via a fast multi-tier loading subsystem (see details in \ref{sec:fast multi-tier loading system}). The inference process initializes the model object and sets the GPU memory address for each tensor. It acquires the base addresses for each GPU (\ie CUDA IPC handles) from the model manager and reads the tensor offset from the tensor index file, facilitating the computation of the tensor GPU memory address (\ie \textit{base + offset}). To ensure the model is fully initialized before inference, the inference process and the model manager perform a synchronization.

\subsection{Multi-Tier Loading Subsystem}\label{sec:fast multi-tier loading system}

To achieve fast and predictable checkpoint loading performance, we design a multi-tier loading subsystem, integrated within the model manager. This subsystem incorporates several techniques:

\mypar{Chunk-based data management} For fast loading performance, we have implemented chunk-based data management with three main features:
(i)~\emph{Utilizing parallel PCIe links}. To mitigate the bottleneck caused by a single PCIe link from storage when loading multiple models into GPUs, we employ parallel DRAM-to-GPU PCIe links to facilitate concurrent checkpoint loading across GPUs.
(ii)~\emph{Supporting application-specific controls}. Our memory pool surpasses simple caching by providing APIs for the allocation and deallocation of memory. This enables fine-grained management of cached or evicted data chunks, based on specific requirements of the application.
(iii)~\emph{Mitigating memory fragmentation}. We address latency and space inefficiencies caused by memory fragmentation by using fixed-size memory chunks.

\mypar{Predictable data path} We have created an efficient data path in our model manager with two main strategies: 
(i)~\emph{Exploiting direct file access}. We use direct file access (\eg `O\_DIRECT' in Linux) to avoid excessive data copying by directly reading data into user space. This method outperforms memory-mapped files (mmap), currently adopted in high-speed loaders such as Safetensors~\cite{safetensors}, which rely on system cache and lack consistent performance guarantees (critical for predictable performance).
(ii)~\emph{Exploiting pinned memory}. We utilize pinned memory to eliminate redundant data copying between DRAM and GPU. This approach allows direct copying to the GPU with minimal CPU involvement, ensuring efficient use of PCIe bandwidth with a single thread.

\mypar{Multi-tier loading pipeline} We have developed a multi-tier loading pipeline to support various storage interfaces and improve loading throughput. This pipeline has three features:
(i)~\emph{Support for multiple storage interfaces}. \sys offers dedicated function calls for various storage interfaces, including local storage (e.g., NVMe, SATA), remote storage (e.g., S3 object store~\cite{aws-s3}), and in-memory storage (pinned memory). It utilizes appropriate methods for efficient data access in each case.
(ii)~\emph{Support for intra-tier concurrency}. To leverage modern storage devices' high concurrency, \sys employs multiple I/O threads for reading data within each storage tier, improving bandwidth utilization.
(iii)~\emph{Flexible pipeline structure}. We use a flexible task queue-based pipeline design, supporting new storage tiers to be efficiently integrated. I/O threads read storage chunks and enqueue their indices (offset and size) for the I/O threads in the next tier.

\section{Efficient Live Migration of LLM Inference}
\label{sec:migration}


In this section, we describe why live migration is the key to effective locality-driven LLM inference, and how to make such a live migration process particularly efficient.






\begin{figure}[t]
    \centering
    \includegraphics[width=\linewidth]{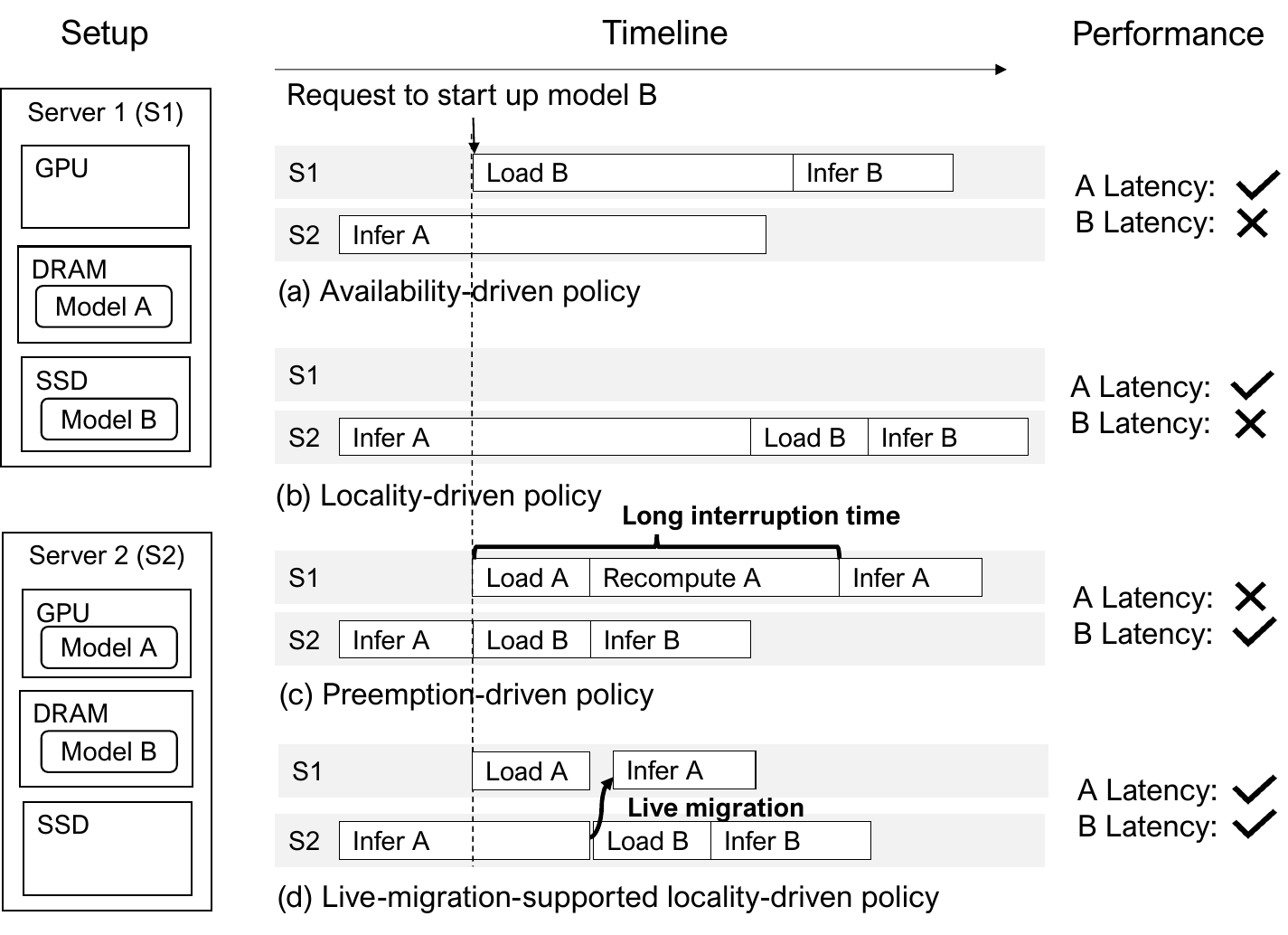}
    \caption{Analysis of different locality-driven policies}
    \label{fig:migration-illustration}
\end{figure}


\subsection{Need for Live Migration} \label{live-migration}

We consider a simple example to analyze the performance of different current approaches in supporting the checkpoint locality. In this example, we have two servers (named Server 1 and Server 2) and two models (named Model A and Model B), as illustrated in Figure~\ref{fig:migration-illustration}. Server 1 currently has Model A in DRAM and Model B in SSD and its GPU is idle, while Server 2 currently has Model B in DRAM, and its GPU is running the inference of Model A.

In Figure~\ref{fig:migration-illustration}, we analyze the performance of potential policies for starting up Model B. Our analysis is based on their impact on the latency performance of both Model A and B: 

\begin{itemize}[leftmargin=*]
    \item \emph{Availability-driven policy} chooses Server 1 currently with an available GPU, and it is agnostic to the location of Model B. As a result, the Model B's startup latency suffers while the Model A remains unaffected.
    \item \emph{Locality-driven policy} opts for the locality in choosing the server and thus launching Model B on Server 2. However, it waits for Model A to complete, making Model B suffer from a long queuing delay. Furthermore, the locality policy leaves Server 1 under-utilized, preventing all servers from being fully utilized.
    \item \emph{Preemption-driven policy} preempts Model A on Server 2 and startups Model B. It identifies that Server 1 is free and reinitiates Model A there. This policy reduces Model B's latency but results in significant downtime for Model A when it performs reloading and recomputation.
    \item \emph{Live-migration-supported locality-driven policy} prioritizes locality without disrupting Model A. It initially preloads Model A on Server 1, maintaining inference operations. When Model A is set on Server 1, its intermediate state is transferred there, continuing the inference seamlessly. Following this, Model B commences on Server 2, taking advantage of locality. This policy optimizes latency for both Models A and B. 
\end{itemize}

According to the examples above, live migration stands out in improving latency for both Model A and Model B among all locality-driven policies.

\subsection{Making Live Migration Efficient}



We aim to achieve efficient live migration of LLM inference, incurring minimal resource overhead and minimal user interruption. We initially considered using the snapshot method from Singularity~\cite{shukla2022singularity}, which involves snapshotting the LLM inference. However, this method is slow due to lengthy snapshot creation and transfer times (\eg typically 10s seconds or even minutes). Dirty-page-based migration might be considered to accelerate virtual machine migration, but this approach is currently not supported in GPU-enabled containers and virtual machines. Hence, we decided to explore live migration methods that can be easily implemented in applications.

To make the live migration method effective for LLM inference, we aim to achieve two objectives: (i) the migrated inference state must be minimal to reduce network traffic, and (ii) the destination server must quickly synchronize with the source server's progress to minimize migration times.

For (i), we propose to migrate tokens (typically 10-100s KB) instead of the large KV-Cache (typically 1-10s GB), as recomputing the KV-Cache based on the migrated tokens on the destination GPU is generally much faster than transferring the dirty state over the network. In certain conditions (\eg given high-bandwidth network and short input sequences), migrating KV-Cache might also be fast yet it still increases cluster network traffic compared to migrating tokens.

For (ii), we leverage an insight from LLM inference: recomputing the KV-Cache for current tokens on the destination GPU is significantly faster (usually an order of magnitude shorter) than generating an equivalent number of new tokens on the source GPU. This approach facilitates efficient convergence of multi-round token-based migration, with the quantity of tokens generated on the source diminishing with each round. For example, time to recompute the KV-Cache for 1000 tokens equals to the time to generate about 100 new tokens according to~\cite{strati2024dejavu}.

\subsection{Multi-Round Live Migration Process}




We implement the above proposal as a multi-round live migration process. In each migration round (step \myc{3}, \myc{4} and \myc{5}), the destination server (referred to as the \emph{dest} server) recomputes the KV cache using the intermediate tokens sent by the source server (referred to as the \emph{src} server). When the gap (\ie the tokens generated after the last round) between the source server and the destination server is close enough, the \emph{src} server stops generating and sends all tokens to the \emph{dest} via the request router, ensuring minimal interruption on ongoing inference during migration. This migration process is depicted in Figure~\ref{fig:seamless migration} with its steps defined below:
\begin{figure}[t]
    \centering
    \includegraphics[width=\linewidth]{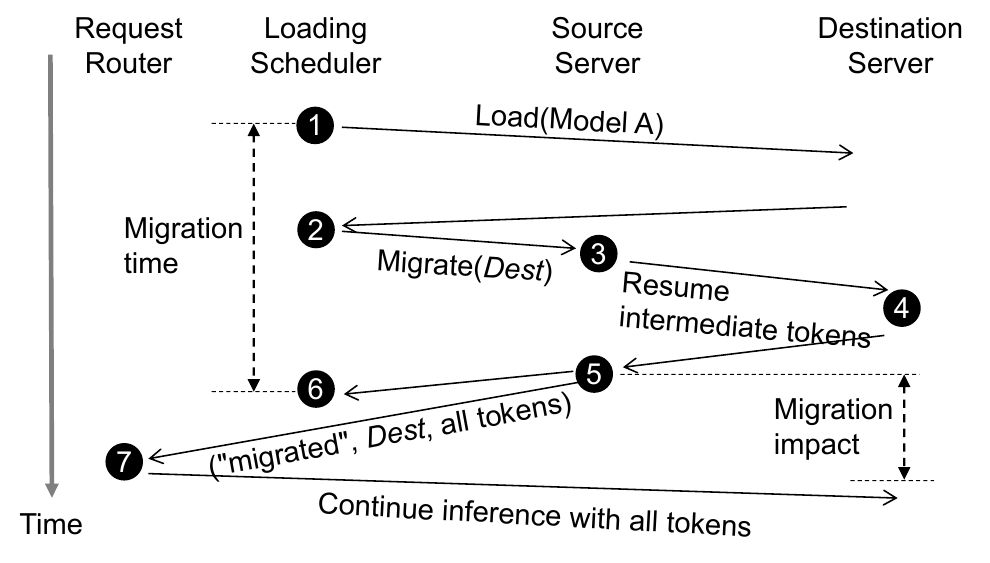}
    \caption{Live migration process for LLM inference}
    \label{fig:seamless migration}
\end{figure}

\begin{enumerate}[leftmargin=*, itemsep=0.5mm, parsep=0.5mm]
    \item The model loading scheduler sends a model loading request to \emph{dest} server 
    to load model A into GPUs. If there is an idle instance of model A on \emph{dest} server, the scheduler skips this step.
    \item After loading, the scheduler sends a migration request carrying the address of \emph{dest} server to \emph{src} server. 
    \item Upon receiving a migrate request, \emph{src} server sets itself as ``migrating'', sends a \empty{resume} request with intermediate tokens (\ie input tokens and the output tokens produced before step 3) to \emph{dest} server if the inference is not completed. Otherwise, it immediately returns to the scheduler. 
    \item \emph{dest} server recomputes KV cache given the tokens in the resume request.
    \item Once \emph{resume} request is done, \emph{src} server stops inference, returns to the scheduler, and replies to the request router with all tokens (\ie the intermediate tokens together with the remaining tokens produced between step 3 and step 5) and a flag ``migrated''. 
    \item The scheduler finishes the migration, unloads model A at 
    \emph{src} server
    and starts loading model B.
    \item The request router checks the flag in the inference response. If it is ``migrated'', the request router replaces \emph{src} server with \emph{dest} server in its route table and sends all tokens to \emph{dest} server to continue inference.
\end{enumerate}

\subsection{Practical Concerns}

\mypar{Handling inference completion} 
The autoregressive nature
of LLM inference may lead to task completion at 
\emph{src} server
between steps \myc{3} and \myc{5}. In such cases, 
\emph{src} server
informs the request router of the inference completion as usual. Additionally, it notifies the loading scheduler, which then instructs 
\emph{dest} server
to cease resuming, terminating the migration.

\mypar{Handling server failures} \sys can manage server failures during LLM inference migration. In scenarios where 
\emph{src} server
fails, if the failure happens during loading (i.e., before step \myc{2} in Figure~\ref{fig:seamless migration}), the scheduler aborts the migration and unloads the model from the destination. If the failure occurs during migration (i.e., between steps \myc{2} and \myc{3}), the scheduler directs the destination to clear any resumed KV cache and unload the model.

In cases where 
\emph{dest} server
fails, if the failure takes place during loading, the migration is canceled by the scheduler. Should the failure occur while resuming, the source notifies the scheduler of the failure and continues with the inference.

\section{Startup-Time-Optimized Model Scheduling} \label{sec:allocation}

\begin{figure}[t]
    \centering
    \includegraphics[width=\linewidth]{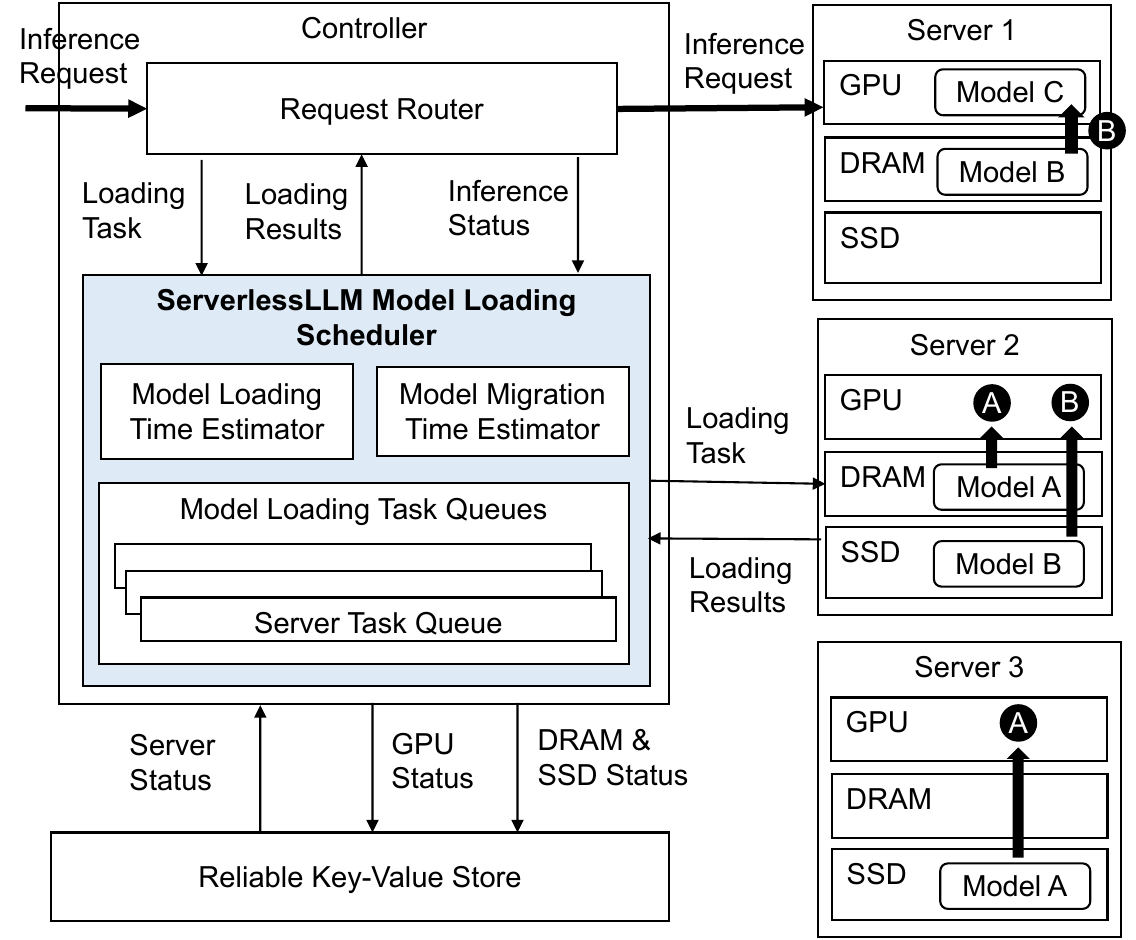}
    \caption{Overview of the model loading scheduler design}
    \label{fig:scheduler-design}
\end{figure}

In this section, we describe the design of the startup-time-optimized model scheduling implemented in \sys's cluster scheduler (denoted as controller), as shown in Figure~\ref{fig:scheduler-design}. This scheduler processes loading tasks from the request router and employs two key components: a model loading time estimator and a model migration time estimator. The former assesses loading times from various storage media, while the latter estimates times for necessary model migrations. For example, as shown in Figure~\ref{fig:scheduler-design}, the scheduler calculates the time to load Model A (indicated by \myc{A}) from different servers' DRAM and SSD, aiding in server selection. Similarly, for Model B (\myc{B}), it assesses whether to migrate Model C to another server or load Model B from Server 2's SSD.

To ensure robust time estimation, the \sys scheduler employs distinct loading task queues for each server, effectively mitigating the impact of contentions caused by concurrent loading activities. Upon assigning a task, it promptly updates the server status—including GPU and DRAM/SSD states—in a reliable key-value store (\eg, etcd~\cite{etcd} and ZooKeeper~\cite{zookeeper}). This mechanism enables \sys to maintain continuity and recover efficiently from failures.

\subsection{Estimating Model Loading Time}

To estimate the time needed to load models from different storage tiers, we consider three primary factors: (i) \emph{queuing time} ($q$), which is the wait time for a model in the server's loading task queue. This occurs when other models are pending load on the same server; (ii) \emph{model size} ($n$), the size of the model in bytes, or its model partition in multi-GPU inference scenarios; (iii) \emph{bandwidth} ($b$), the available speed for transferring the model from storage to GPUs. \sys tracks bandwidth for network, SSD, and DRAM, allowing us to calculate loading time as $q + n / b$. Here, $q$ accumulates from previous estimations for the models already in the queue.

For precise estimations, we have implemented: (i) Sequential model loading per server, with single I/O queues for both Remote-SSD and SSD-DRAM paths (since these paths are shared by multiple GPUs on a server), reducing bandwidth contention which complicates estimation; (ii) In multi-tier storage, \sys uses the slowest bandwidth for estimation because of \sys's pipeline loading design. For example, when SSD and DRAM are both involved, SSD bandwidth is the critical bottleneck since it is orders of magnitude slower than DRAM; (iii) The scheduler monitors the loading latency returned by the servers. It leverages the monitoring metrics to continuously improve its estimation of the bandwidth through different storage media. 

\subsection{Estimating Model Migration Time}

For live migration time estimation, our focus is on model resuming time (as shown in step \myc{4} in Figure~\ref{fig:seamless migration}), as this is significantly slower (seconds) than token transfer over the network (milliseconds). We calculate model resuming time considering: (i) \emph{input tokens} ($t_{in}$), the number of tokens in the LLM's input prompt; (ii) \emph{output tokens} ($t_{out}$), the tokens generated so far; and (iii) \emph{model-specific parameters} ($a$ and $b$), which vary with each LLM's batch sizes and other factors, based on LLM system studies like vLLM~\cite{vllm}. With all the above factors, we can compute the model resuming time as $a \times (t_{in} + t_{out}) + b$.

However, obtaining real-time output tokens from servers for the scheduler can lead to bottlenecks due to excessive server interactions. To circumvent this, we developed a method where the scheduler queries the local request router for the inference status of a model, as illustrated in Figure~\ref{fig:scheduler-design}. With the inference duration ($d$) and the average time to produce a token ($t$), we calculate $t_{out} = d / t$.

For selecting the optimal server for model migration, \sys employs a dynamic programming approach to minimize migration time. 

\subsection{Practical Concerns}

\mypar{Selecting best servers} Utilizing our time estimations, \sys evaluates all servers for loading the forthcoming model, selecting the one offering the lowest estimated startup time. The selection includes the server ID and GPU slots to assign. If no GPUs are available, even after considering migration, the loading task is held pending and retried once the request router informs the scheduler to release GPUs.

\mypar{Handling scheduler failures} \sys is built to withstand failures, utilizing a reliable key-value store to track server statuses. On receiving a server loading task, its GPU status is promptly updated in this store. Post server's confirmation of task completion, the scheduler updates the server's storage status in the store. Once recorded, the scheduler notifies the request router of the completion, enabling request routing to the server. In the event of a scheduler failure, recovery involves retrieving the latest server status from the key-value store and synchronizing it across all servers.

\mypar{Scaling schedulers} 
The performance of the loading scheduler has been significantly enhanced by implementing asynchronous operations for server status reads, writes, and estimations. Current benchmarks demonstrate its capability to handle thousands of loading tasks per second on a standard server. Plans for its distributed scaling are earmarked for future development.

\mypar{Resource fairness} \sys treats all models with equal importance and it ensures migrations do not impact latency. While we currently adopt sequential model loading on the I/O path, exploring concurrent loading on servers with a fairness guarantee is planned for future work.

\mypar{Estimator accuracy} Our estimator can continuously improve their estimation based on the monitored loading metrics returned by the servers. They offer sufficient accuracy for server selection, as shown in Section~\ref{evaluation}.

\begin{figure*}[t]

\centering
 \includegraphics[width=0.4\linewidth, keepaspectratio]{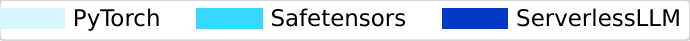}
\\
\begin{subfigure}{0.60\textwidth}
    \includegraphics[height=3.5cm, keepaspectratio]{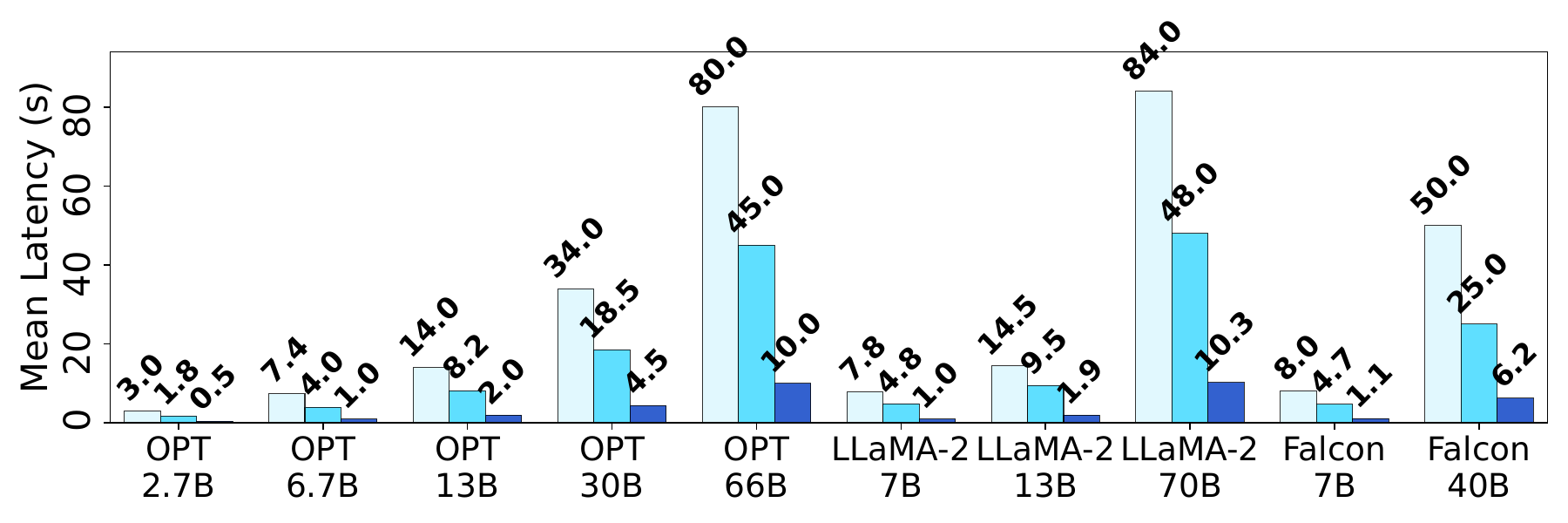}
    \caption{Checkpoint loading speed.}
    \label{fig:loading speed vs loader}
  \end{subfigure}%
  \begin{subfigure}{0.40\textwidth}

    \includegraphics[height=3.5cm, keepaspectratio]{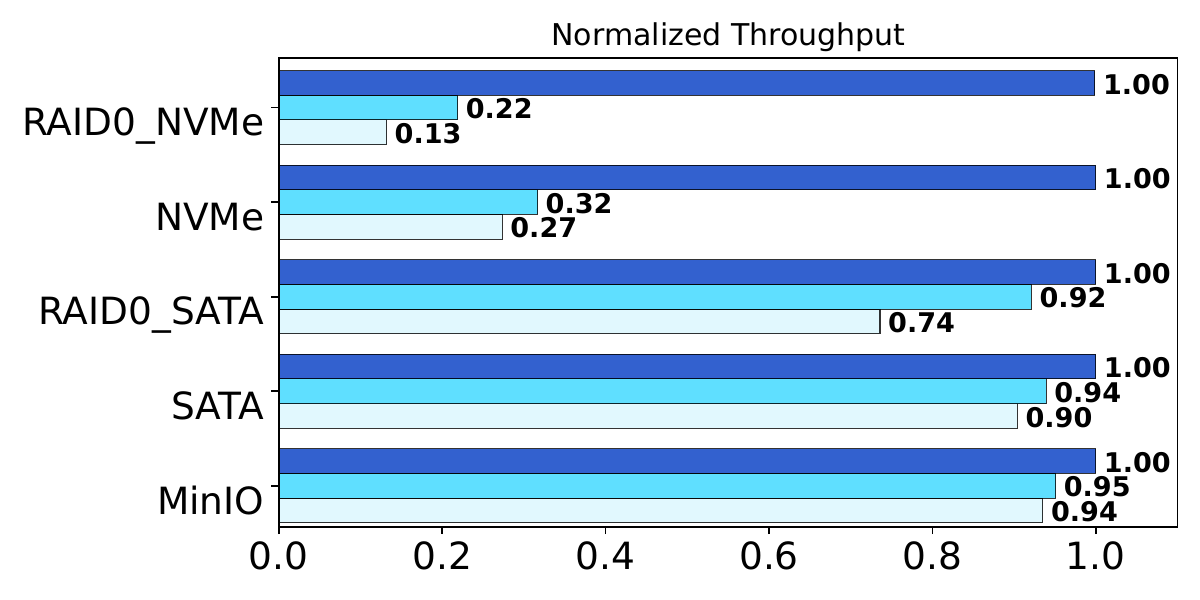}
    \caption{Normalized bandwidth utilization.}
    \label{fig:bandwidth utilization}
  \end{subfigure}
  \caption{Checkpoint loading performance.}
\end{figure*}

\section{Evaluation} \label{evaluation}

This section offers a comprehensive evaluation of \sys, covering three key aspects: (i) assessing the performance of our loading-optimized checkpoints and model manager, (ii) examining the efficiency and overheads associated with live migration for LLM inference, and (iii) evaluating \sys against a large-scale serverless workload, modelled on real-world serverless trace data. 

\subsection{Evaluation Setup}
\mypar{Setup}
We have two test beds: (i)~a GPU server has 8 NVIDIA A5000 GPUs, 1TB DDR4 memory and 2 AMD EPYC 7453 CPUs, two PCIe 4.0-capable NVMe 4TB SSDs (in RAID 0) and two SATA 3.0 4TB SSDs (in RAID 0). 
This server is connected to a storage server via 1 Gbps networks on which we have deployed MinIO~\cite{minio}, an S3 compatible object store;
(ii)~a GPU cluster with 4 servers connected with 10 Gbps Ethernet connections. Each server has 4 A40 GPUs, 512 GB DDR4 memory, 2 Intel Xeon Silver 4314 CPUs and one PCIe 4.0 NVMe 2TB SSD.


\mypar{Models}
We use state-of-the-art LLMs, including
OPT~\cite{DBLP:journals/corr/abs-2205-01068}, LLaMA-2~\cite{touvron2023llama2} and Falcon~\cite{falcon40b} in different sizes. For cluster evaluation (\S\ref{sec:evaluation_scheduler} and \S\ref{sec:evaluation_end2end}) on test bed (ii), following prior work~\cite{DBLP:conf/osdi/LiZZL00HCZGS23}, we replicate OPT-6.7B/OPT-13B/OPT-30B models for 32/16/8 instances respectively (unless otherwise indicated) that are treated as different models during evaluation.


\mypar{Datasets}
We use real-world LLM datasets as the input to models.
This includes GSM8K~\cite{DBLP:journals/corr/abs-2110-14168} that contains problems created by human problem writers, and ShareGPT~\cite{sharegpt4} that contains multilanguage chat from GPT4.
Since the models we used can handle at most 2048 context lengths, we truncate the input number of tokens to the max length.
We also randomly sample 4K samples from each dataset to create a mixed workload, emulating real-world inference workloads.


\mypar{Workloads} Since there are no publicly available LLM serverless inference workloads, we use Azure Serverless Trace~\cite{254430} which is a representative serverless workload used in recent serverless studies~\cite{roy2022icebreaker} and model-serving studies~\cite{DBLP:conf/osdi/LiZZL00HCZGS23, shepherd}. We designate functions to models and creates bursty request traces (CV=8 using Gamma distribution), following the workload generation method used in AlpaServe~\cite{DBLP:conf/osdi/LiZZL00HCZGS23}. We then scale this trace to the desired requests per second (RPS). 
For cluster evaluation, we replicate each model based on its popularity and distribute them across nodes' SSDs using round-robin placement until the total cluster-wide storage limit is reached. Optimization of checkpoint placement is considered a separate issue and is not addressed in this paper.
For all experiments (unless we indicate otherwise), we report the model startup latency, a critical metric for serverless inference scenarios. When migration or preemption is enabled, this latency is added with pause latency, accounting for the impacts of delays.

\subsection{\sys Checkpoint Loading}\label{sec:evaluation_checkpoint}
We now evaluate the model manager's effectiveness in reducing the 
model loading
latency. For our experiments, we test the checkpoint read on test bed (i). We record reads from 20 copies of each model checkpoint to get a statistically significant performance report. We clear the page and inode caches after checkpoint copies are made to ensure a cold start. For each type of model, we randomly access the 20 copies 
to simulate real-world access patterns.


\mypar{Loading performance} 
We aim to quantify the performance gains achieved by the \sys checkpoint manager.
We compare PyTorch~\cite{pytorch} and Safetensors~\cite{safetensors}, representing the read-by-tensor checkpoint loading and mmap-based checkpoint loading, respectively. We use all types of models with all checkpoints in FP16 and run the test on RAID0-NVMe SSD having a throughput of 12 GB/s. 


Figure~\ref{fig:loading speed vs loader} shows the performance comparison in terms of mean latency for all the models\footnote{The number after the model name represents the number of parameters in the figure and B stands for Billion.}. We observe that \sys is 6X and 3.6X faster than PyTorch and Safetensors, respectively, for our smallest model (OPT-2.7B). We observe similar results with the largest model (LLaMA-2-70B) where \sys is faster than PyTorch and Safetensors by 8.2X and 4.7X respectively. Safetensors is slower than \sys due to a lot of page faults (112K for LLaMA-2-7B) on cold start. In contrast, \sys's checkpoint manager leverages direct I/O and realizes chunk-based parallel loading, all contributing to the significant improvement in loading throughput.
PyTorch is about 2X slower than Safetensors in our evaluation, consistent with the results in a public benchmark~\cite{safetensors-report} reported by Safetensors. The primary reason is that PyTorch first copies data into host memory and then into GPU memory.

Furthermore, we observe that the loading performance of \sys is agnostic to the type of the model. For example, the performance of both OPT-13B and LLaMA-2-13B is similar signifying the fact that the performance is only dependent on the checkpoint size. 

\mypar{Loading performance with LoRA adapters}
\sys also supports loading LoRA adapters~\cite{hu2021lora} in PEFT format~\cite{peft}. We conducted experiments using the same setting in~\cite{sheng2023s}. For an adapter (rank=32, size=1GB) of LLaMA-70B model, \sys achieves 83.5ms loading latency which is 4.4X faster than Safetensors whose loading latency is 370ms. This demonstrates \sys's loader design efficiency in small checkpoint loading.



\mypar{Harness full bandwidth of the storage devices} We now move to understand if \sys can utilize the entire bandwidth that a storage medium offers to achieve low latency. We use the same setup as described above. We choose LLaMA-2-7B to represent the SOTA LLM model. We use FIO~\cite{Axboe_Flexible_I_O_Tester_2022} with the configuration of asynchronous 4M direct sequential read with the depth of 32 as the optimal baseline and optimized throughput using the result in all storage media. We test various settings of FIO to make sure the configuration chosen has the highest bandwidth on each storage media.
For object storage over the network, we use the official MinIO benchmark to get the maximum throughput.


Figure~\ref{fig:bandwidth utilization} shows the bandwidth utilization across different storage devices, normalized relative to the measurements obtained using FIO and MinIO.
The storage device from bottom to top is ascending in maximum bandwidth. We observe that \sys's model manager is capable of harnessing different storage mediums and saturating their entire bandwidth to get maximum performance. Interestingly, we observe that \sys is well suited for faster storage devices such as RAID0-NVMe compared to Pytorch and Safetensors. It shows that existing mechanisms are not adaptive to newer and faster storage technology. Despite the loading process passing through the entire memory hierarchy, \sys is capable of saturating the bandwidth highlighting the effectiveness of pipelining the loading process.





\begin{figure}[t]
    \centering
    \includegraphics[width=1.0\linewidth]{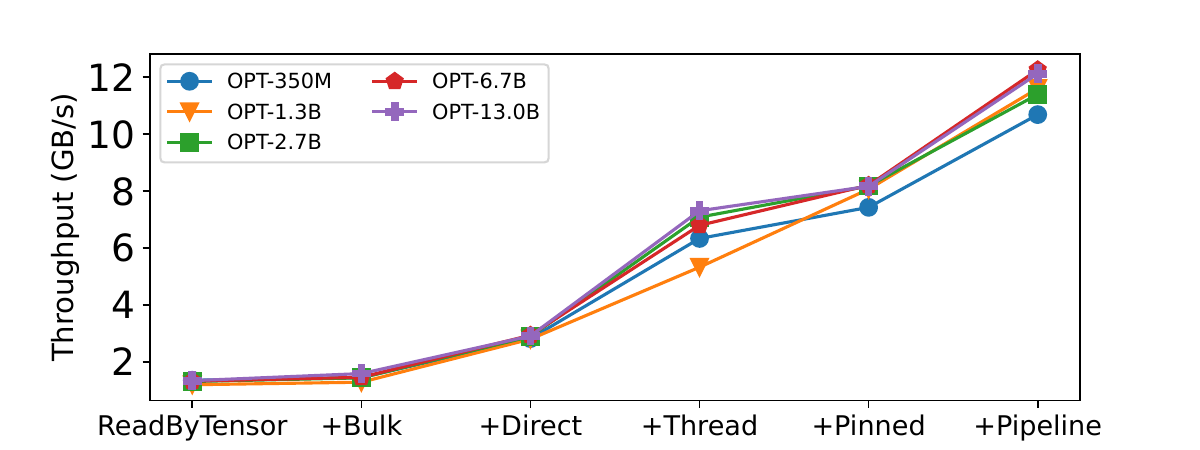}
    \caption{Performance breakdown of checkpoint loaders.}
    \label{fig:checkpoint loader breakdown}
\end{figure}

\mypar{Performance breakdown} We now move to highlight how each optimization within the model manager contributes towards the overall performance. We run an experiment using RAID0-NVMe with various OPT models. We start from the basic implementation (ReadByTensor) and incrementally add optimizations until the Pipeline implementation. Figure ~\ref{fig:checkpoint loader breakdown} shows the performance breakdown for each model. We observe similar contributions by different optimizations for all the models despite having different checkpoint size. 

Bulk reading improves 1.2x throughput, mitigating the throughput degradation from reading small tensors one after another (on average one-third of the tensors in the model are less than 1MB). Direct IO improves 2.1x throughput, bypassing cache and data copy in the kernel. Multi-thread improves 2.3x throughput, as multiple channels within the SSD can be concurrently accessed. Pinned memory provides a further 1.4x throughput, bypassing the CPU with GPU DMA. Pipeline provides a final 1.5x improvement in throughput, helping to avoid synchronization for all data on each storage tier.

We run \sys in a container to limit the CPU cores it can use. 
We find that with 4 CPU cores, \sys can achieve maximum bandwidth utilization.
We set a sufficiently large chunk size in bulk reading (16MB) to involve less number of reads and also pinned memory-based chunk pool does not need extra CPU cycles for data copy.

\subsection{\sys Model Scheduler}\label{sec:evaluation_scheduler}

\begin{figure}[t]

\begin{subfigure}[t]{\linewidth}
 \centering
\includegraphics[width=0.8\linewidth]{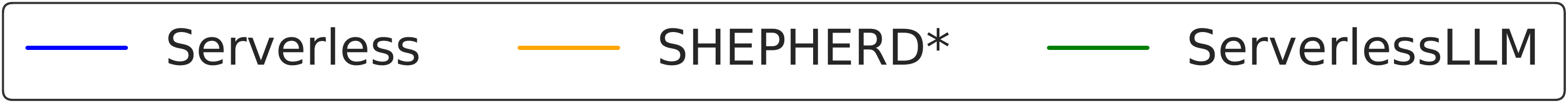}     
 \end{subfigure} \\
    \centering
    \begin{subfigure}[b]{0.32\linewidth}
        \includegraphics[width=\linewidth]{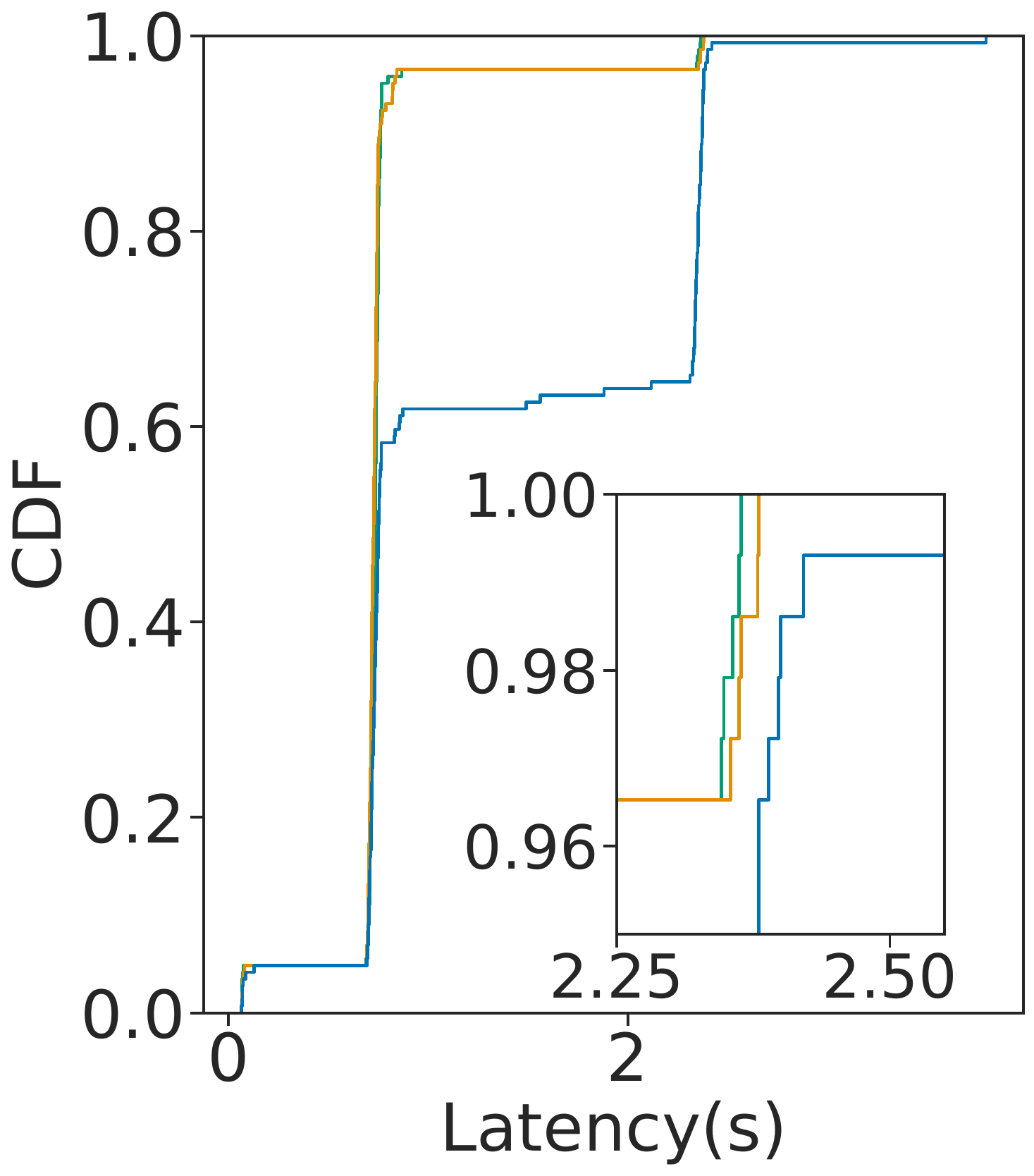}
        \caption{GSM8K, RPS=0.2}
        \label{fig:cdf-gsm8l-0.2}
    \end{subfigure}
    \hfill 
    \begin{subfigure}[b]{0.32\linewidth}
        \includegraphics[width=0.95\linewidth]{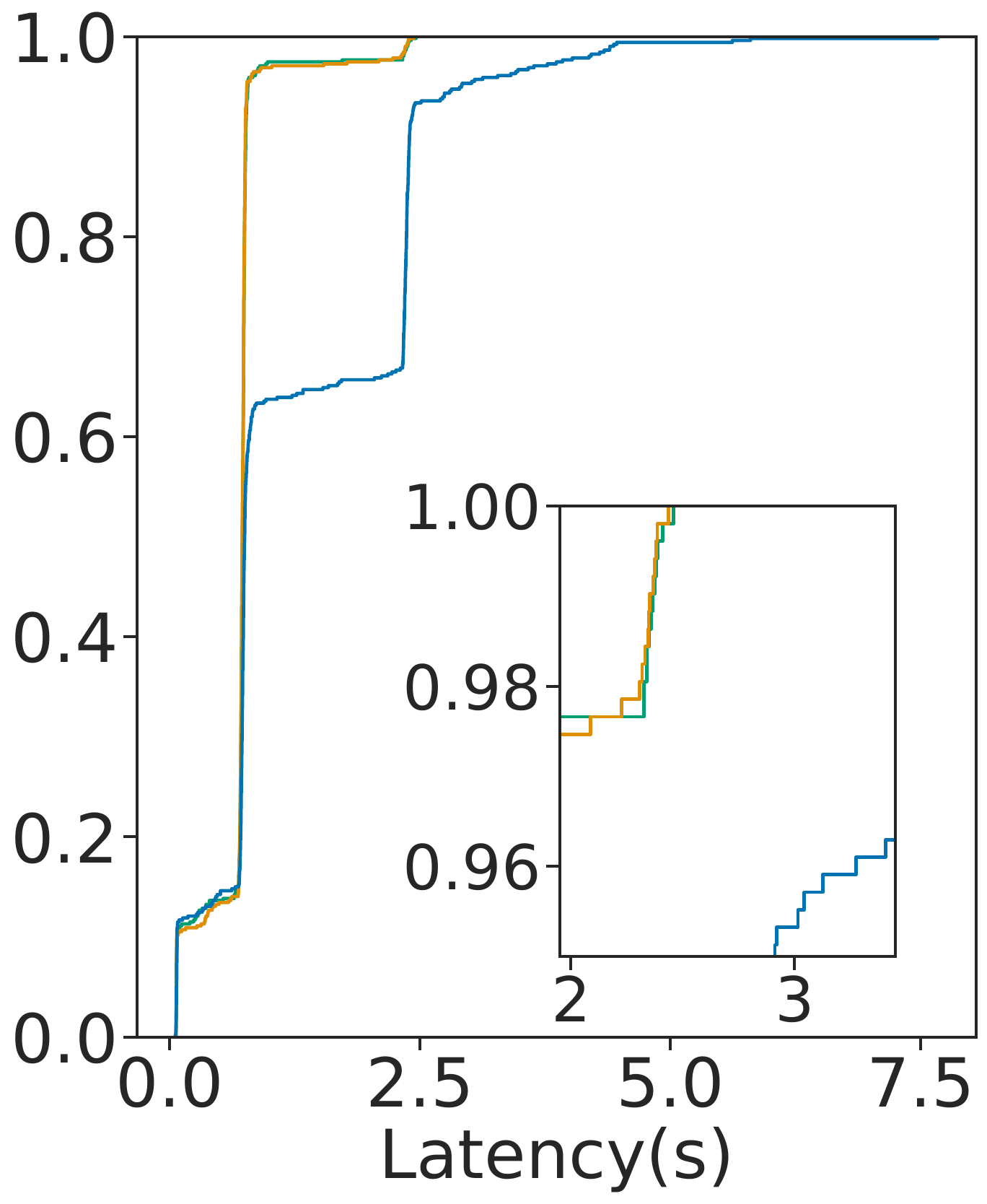}
        \caption{GSM8K, RPS=0.8}
        \label{fig:cdf-gsm8k-0.8}
    \end{subfigure}
    \hfill 
    \begin{subfigure}[b]{0.32\linewidth}
        \includegraphics[width=0.95\linewidth]{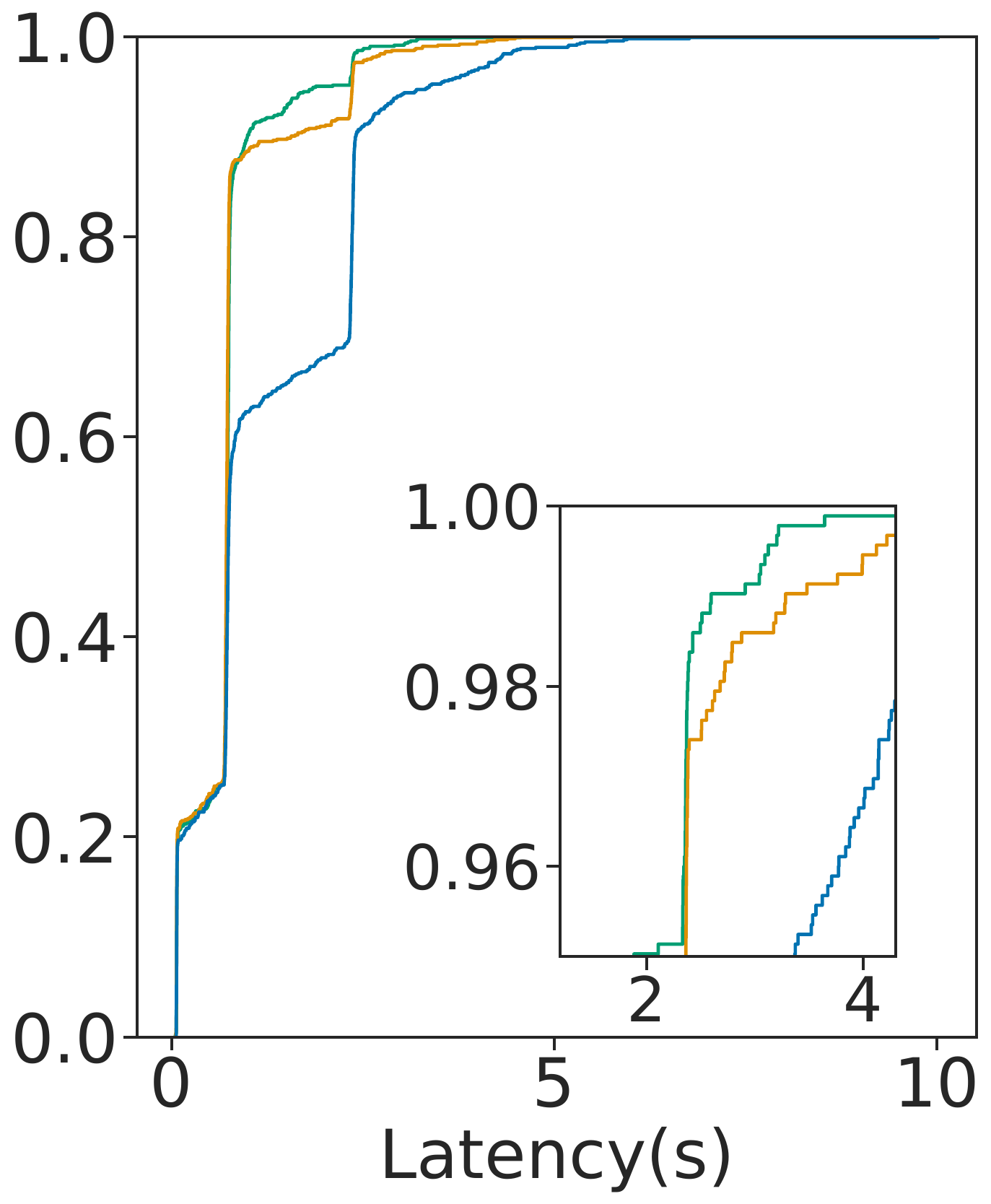}
        \caption{GSM8K, RPS=1.4}
        \label{fig:cdf-gsm8k-1.6}
    \end{subfigure}\\
    \begin{subfigure}[b]{0.32\linewidth}
        \includegraphics[width=\linewidth]{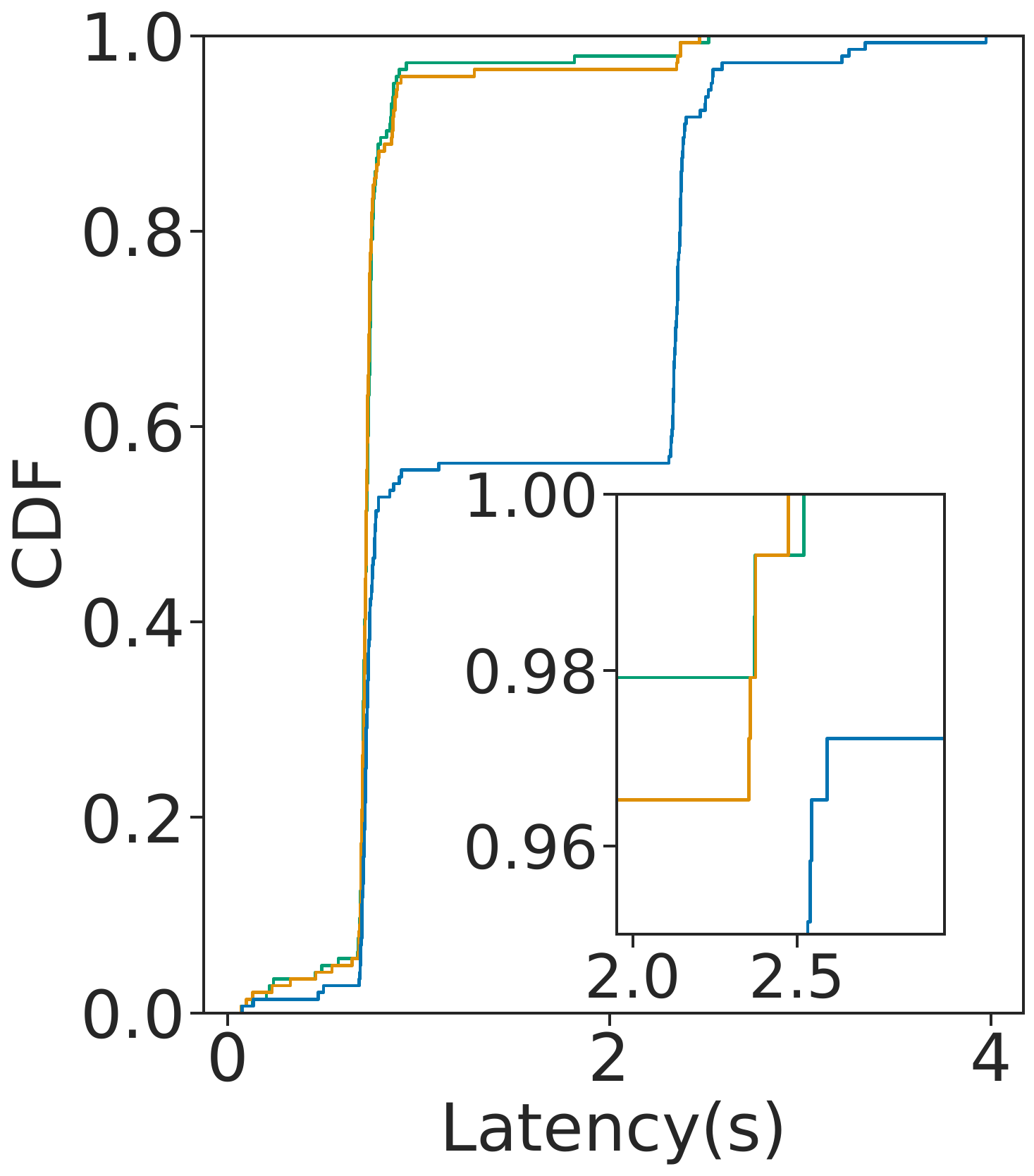}
        \caption{ShareGPT, RPS=0.2}
        \label{fig:cdf-sharegpt-0.2}
    \end{subfigure}
    \hfill 
    \begin{subfigure}[b]{0.32\linewidth}
        \includegraphics[width=0.95\linewidth]{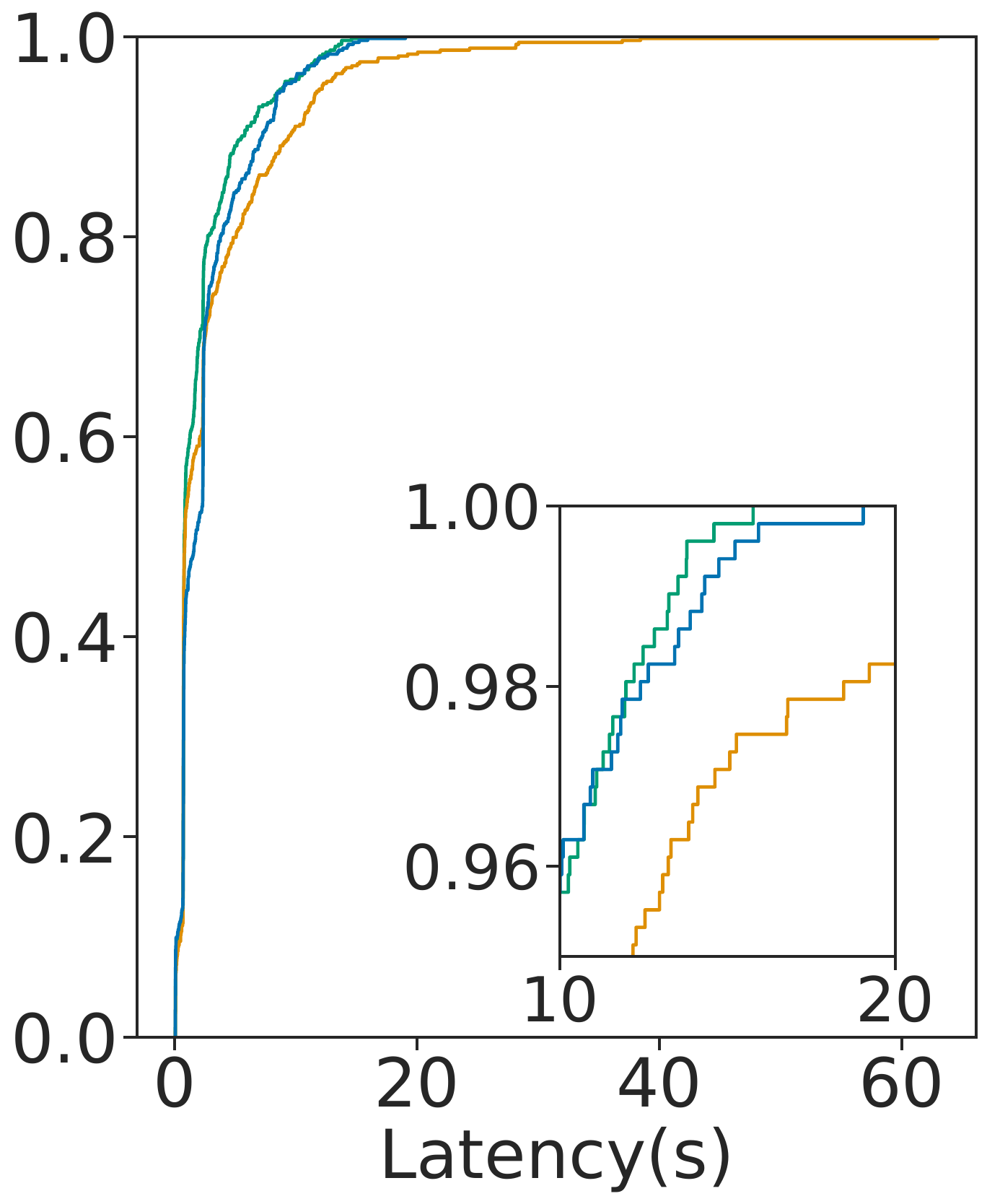}
        \caption{ShareGPT, RPS=0.8}
        \label{fig:cdf-sharegpt-0.8}
    \end{subfigure}
    \hfill 
    \begin{subfigure}[b]{0.32\linewidth}
        \includegraphics[width=0.95\linewidth]{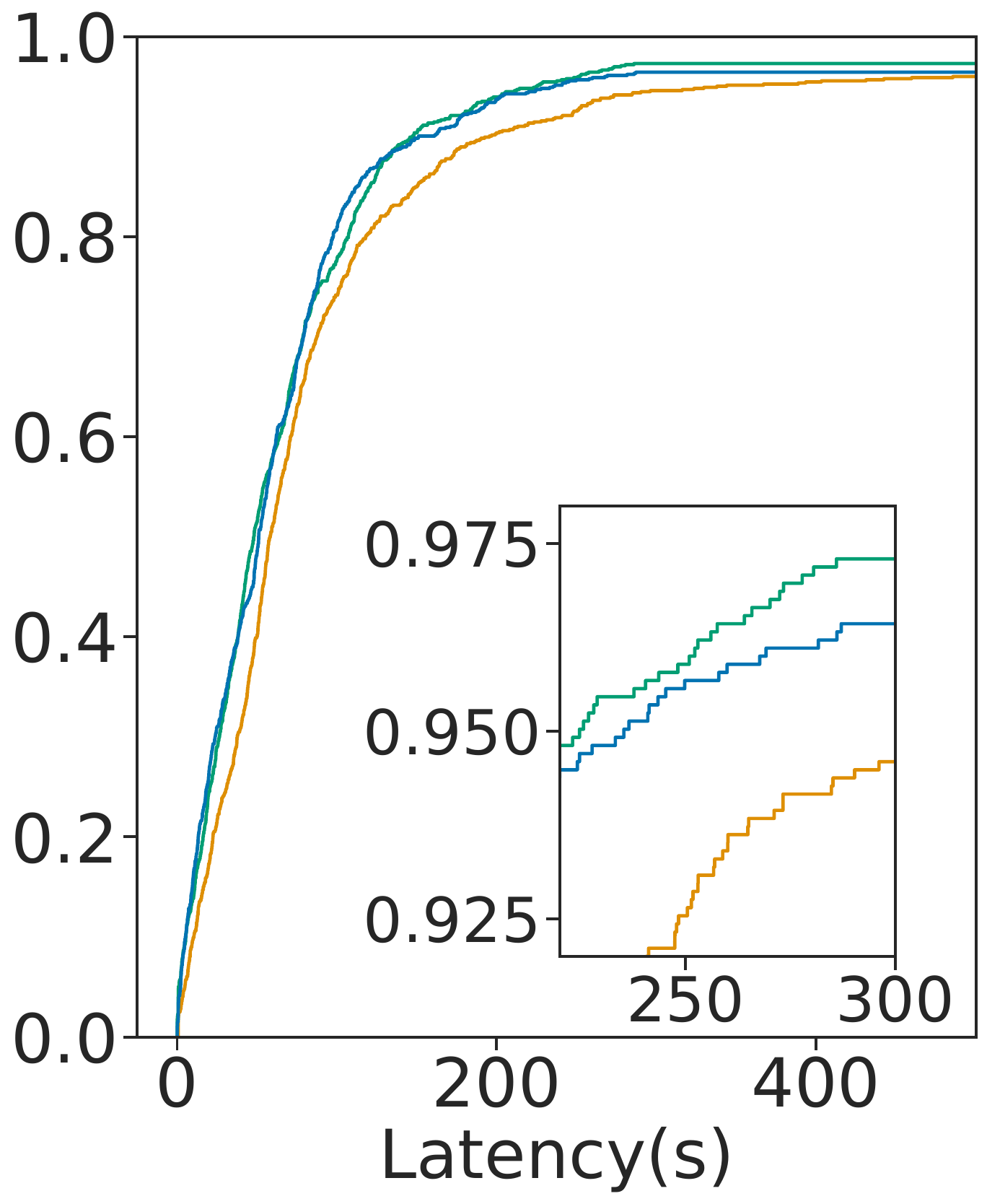}
        \caption{ShareGPT, RPS=1.4}
        \label{fig:cdf-sharegpt-1.4}
    \end{subfigure}
    \caption{Impacts of RPS on model loading schedulers. }
    \label{fig:scheduler_latency_vs_rps}
\end{figure}

In this section, we evaluate the performance of the \sys's cluster scheduler on test bed (ii). We compare \sys against two schedulers -- the de-facto serverless scheduler and Shepherd~\cite{shepherd} scheduler. The serverless scheduler randomly chooses any GPU available and does not comprise any optimization for loading time. We implement Shepherd scheduler and use \sys's loading time estimation strategy to identify the correct GPU. We call the modified scheduler as Shepherd*. Therefore, in principle, Shepherd* and \sys will choose the same GPU. However, Shepherd* will continue to rely on preemption, while \sys will rely on live migration to ensure lower latency times. 

Figure~\ref{fig:cdf-gsm8l-0.2} shows the result of a scenario where we run all three schedulers against OPT-6.7B model and GSM8K and ShareGPT dataset while increasing the requests per second. ShareGPT dataset's average inference time is 3.7X longer than GSM8K. Figure~\ref{fig:cdf-gsm8l-0.2} and Figure~\ref{fig:cdf-sharegpt-0.2} show the case where there is no locality contention for both datasets. The serverless scheduler cannot take advantage of locality-aware scheduling unlike \sys and Shepherd* leading to longer latency. For 40\% of the time, the model is loaded from SSD due to random allocation of the GPUs. As there is no migration or preemption, the performance of Shepherd and \sys is similar. 

When the schedulers are subjected to medium requests per second, for GSM8K (Figure~\ref{fig:cdf-gsm8k-0.8}, without locality-aware scheduling, the loading times start causing queueing latency leading with Serverless scheduler resulting in increasing the P99 latency by 1.86X. As there is no migration or preemption, the performance of Shepherd and \sys is similar. 
With a longer inference time with ShareGPT (Figure~\ref{fig:cdf-sharegpt-0.8}, we even observe 2X higher P99 latency with Shepherd* compared to \sys due to preemption. As \sys relies on live migration in case of locality contention, \sys performs better than the other schedulers despite the number of migrations is higher (114 out of 513 total requests) than the number of preemptions (40 out of 513 total requests). 

On further stressing the system by increasing the requests per second to 1.4, for GSM8K, one can clearly observe the impact of live migration and preemption. \sys outperforms Shepherd* and Serverless schedulers by 1.27X and 1.95X on P99 latency respectively. There are 9 preemptions and 53 migrations respectively for a total of 925 requests. As discussed in Section~\ref{live-migration}, preemptions lead to longer latency compared to migrations. We also observe that with Shepherd*, 
model checkpoints are
read from SSD 2X times more than with \sys. With ShareGPT (figure~\ref{fig:cdf-sharegpt-1.4}, we observe that the GPU 
occupancy reaches 100\% leading to requests timeouts with all the three schedulers\footnote{Based on the average inference time of OPT-6.7B on ShareGPT dataset, the maximum theoretically RPS is 1.79.}. Shepherd behaves the worst compared to Serverless and \sys schedulers, \ie 1.43X and 1.5X higher P95 latency respectively. \sys and Shepherd* issue 64 migrations and 166 preemptions, respectively for a total of 925 requests. In this scenario, \sys's effectiveness is constrained by resource limitations.

\begin{figure}[t]
\begin{subfigure}[b]{\linewidth}
 \centering
\includegraphics[width=0.8\linewidth]{figs//evaluation/legend-serverlessllm.pdf}     
 \end{subfigure} \\
  \centering
  \begin{subfigure}{0.49\linewidth}
    \includegraphics[width=0.9\linewidth]{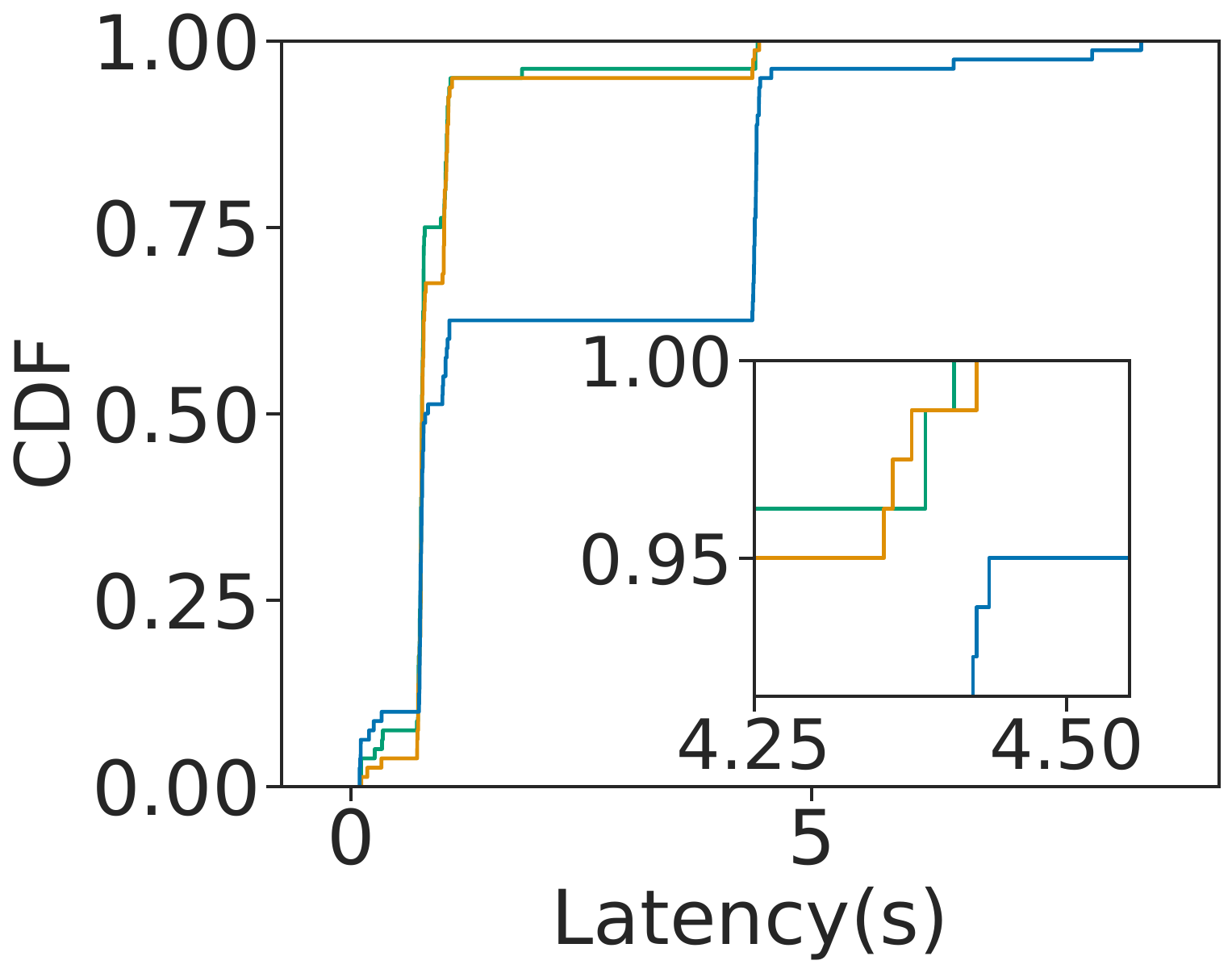}
    \caption{OPT-13B GSM8K}
    \label{fig:sub1}
  \end{subfigure}
  \hfill 
  \begin{subfigure}[b]{0.49\linewidth}
    \includegraphics[width=0.85\linewidth]{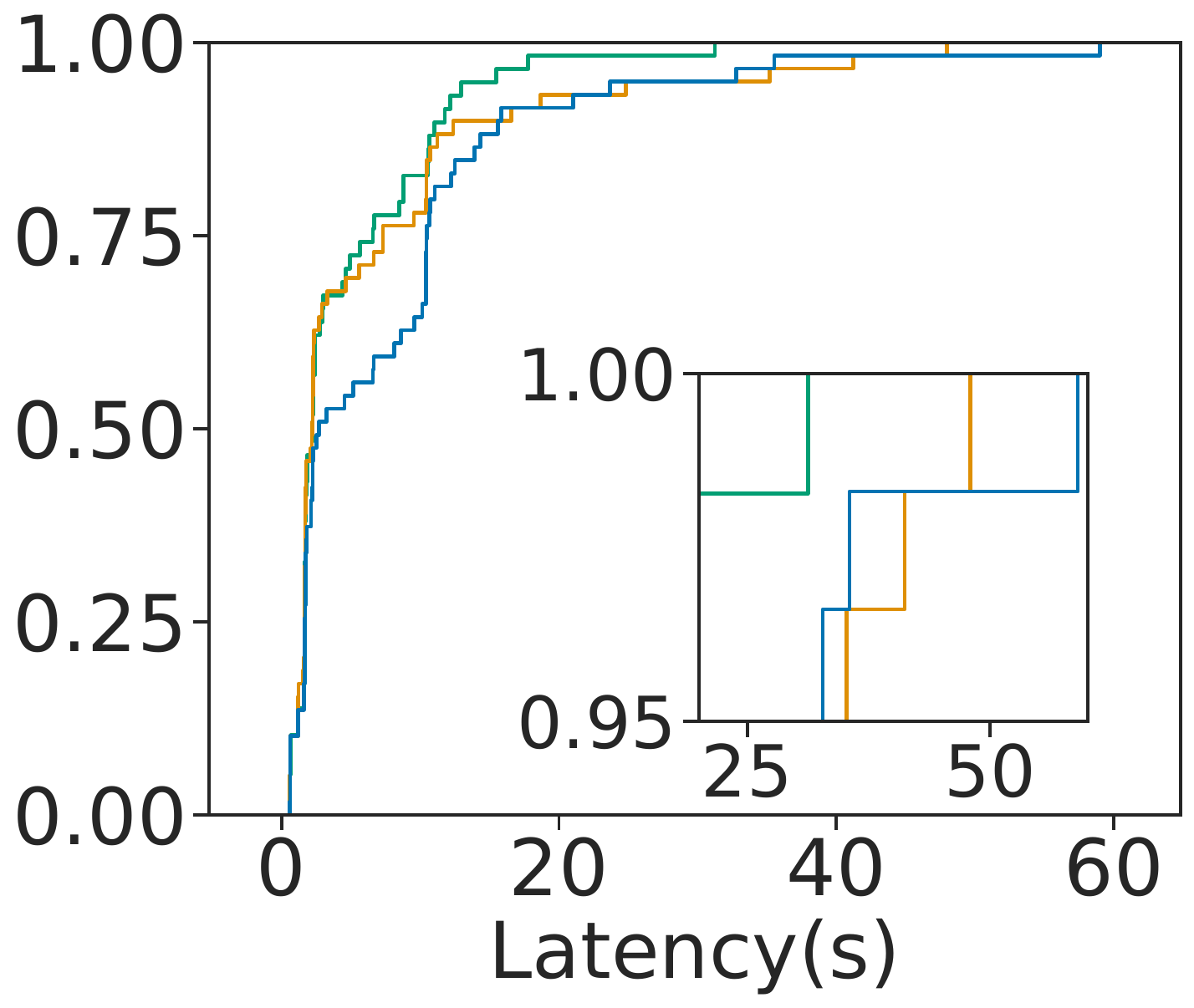}
    \caption{OPT-30B GSM8K}
    \label{fig:sub2}
  \end{subfigure}
  \begin{subfigure}[b]{0.49\linewidth}
    \includegraphics[width=0.9\linewidth]{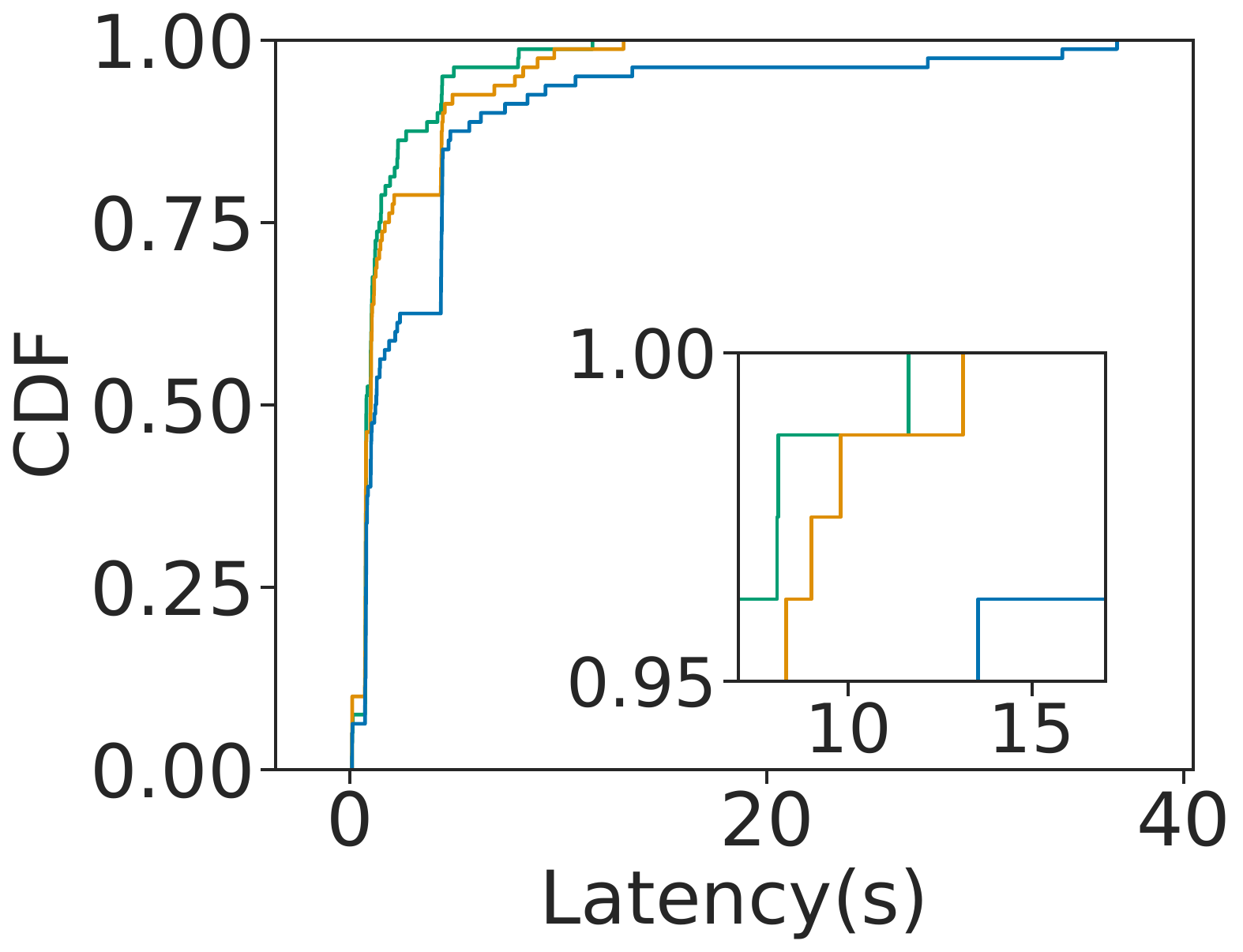}
    \caption{OPT-13B ShareGPT}
    \label{fig:sub3}
  \end{subfigure}
  \hfill
  \begin{subfigure}[b]{0.49\linewidth}
    \includegraphics[width=0.85\linewidth]{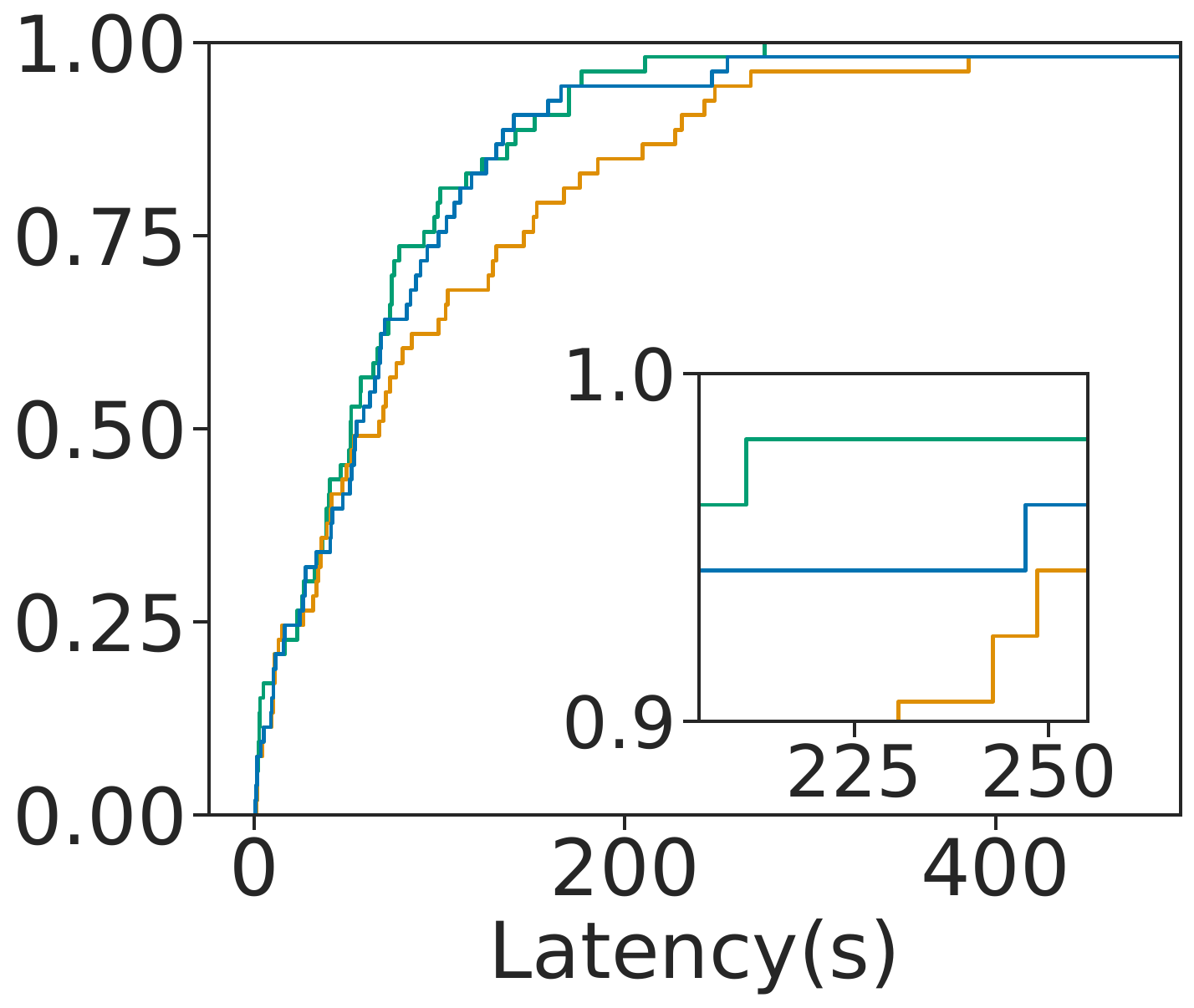}
    \caption{OPT-30B ShareGPT}
    \label{fig:sub4}
  \end{subfigure}
  \caption{Impacts of datasets and models on model loading schedulers.}
  \label{fig:scheduler_latency_vs_model_size}
\end{figure}

We further stress the system by running even larger models (OPT-13B and OPT-30B) with GSM8K and ShareGPT datasets. Figure~\ref{fig:scheduler_latency_vs_model_size} shows the results for those experiments. locality-aware scheduling is more important for larger models as caching them in the main memory can reap better performance. As \sys and Shepherd* are both locality-aware, they can make better decisions while scheduling the requests leading to better performance. As Serverless scheduler makes decisions randomly, for GSM8K, we observe that for 35-40\% times, the model is loaded from SSD leading to poor performance. We see similar behavior for ShareGPT, OPT-13B experiment too. For the OPT-30B ShareGPT case, the model size is 66 GB. Hence, only two models can be stored in the main memory at any given time reducing the impact of locality-aware scheduling. 
Even in this extreme case, \sys still achieves 35\%  and 45\% lower P99 latency compared to Serverless and Shepherd* respectively.




\mypar{Time Estimation}
The GPU time estimation error is bounded at 5ms, while the SSD loading error is bounded at 40ms. However, we do observe instability in CUDA driver calls. For instance, when migrating a model, we noted that cleaning up GPU states (e.g., KV cache) using \texttt{torch.cuda.empty\_cache()} can lead to inaccurate estimations, resulting in an average underestimation of 25.78 ms. While infrequent, we observed a maximum underestimation of 623 ms during GPU state cleanup in one out of 119 migrations (as depicted in Figure~\ref{fig:cdf-sharegpt-0.8}).




\subsection{Entire \sys in Action}\label{sec:evaluation_end2end}

We aimed to deploy the entire \sys with a serverless workload on test bed (ii). Here, we compare \sys against state-of-the-art distributed model serving systems: (i)~Ray Serve (Version 2.7.0), a version we 
have extended
to support serverless inference scenarios with performance that can match SOTA serverless solutions such as KServe; (ii)~Ray Serve with Cache, a version we improved to adopt a local 
SSD cache on each server (utilizing the LRU policy as in \sys) to avoid costly model downloads; and (iii)~KServe (Version 0.10.2), the SOTA serverless inference system designed for Kubernetes clusters. 

For best performance, Ray Serve and its cache variant are both 
enhanced by
storing model checkpoints on local SSDs and estimating download latency by assuming 
an exclusively occupied 10 Gbps network.
For each system, we set the maximum concurrency to one and set the keep-alive period equal to its loading latency, following prior work~\cite{romero2021infaas}.
We launch parallel LLM inference clients to generate various workloads, where each request has a timeout threshold of 300 seconds. 

\begin{figure}[t]
    \centering
    \begin{subfigure}{0.49\textwidth}
    \centering
    \includegraphics[width=0.8\linewidth]{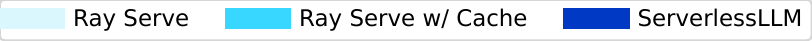}
    \end{subfigure}%
    \\
    \centering
    \begin{subfigure}{0.49\linewidth}
    \includegraphics[height=3.2cm, keepaspectratio]{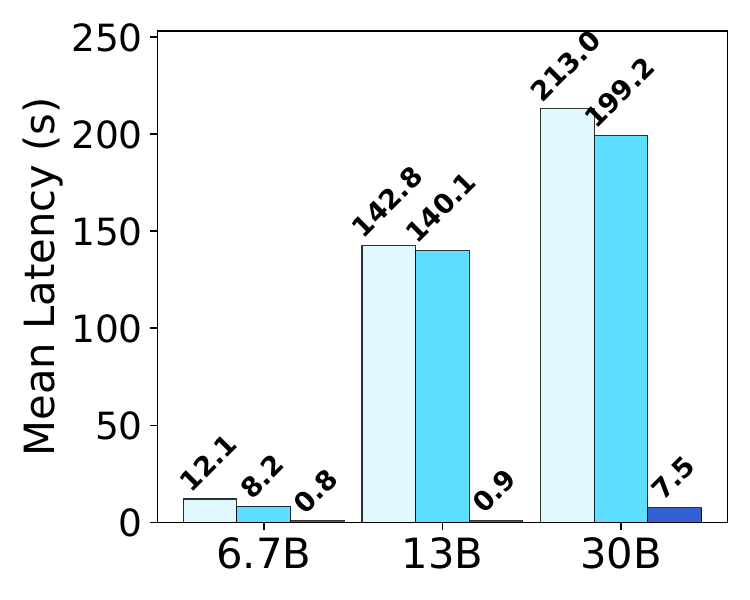}
    \vspace{-0.10in}\
    \caption{GSM8K}
    \label{fig:latency vs gsm8k}
    \vspace{-0.15in}\
    \end{subfigure}%
    \hfill
    \begin{subfigure}{0.49\linewidth}
    \includegraphics[height=3.24cm, keepaspectratio]{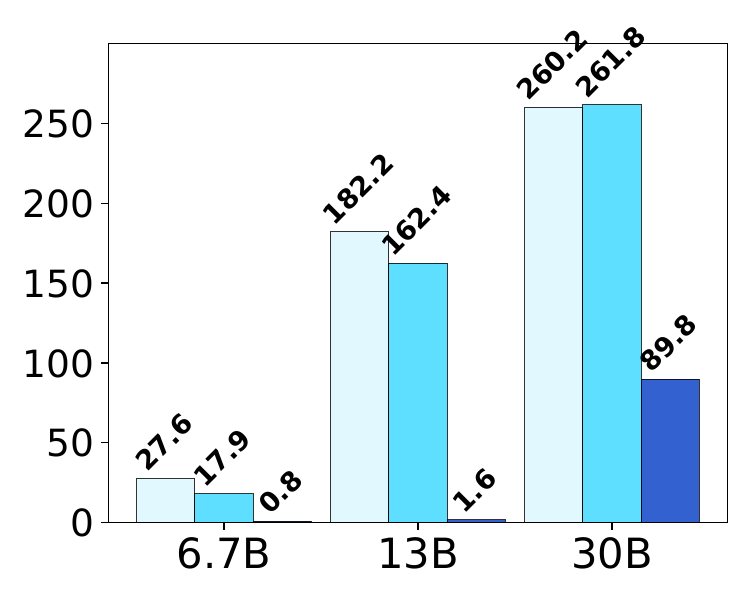}
    \vspace{-0.10in}\
    \caption{ShareGPT}
    \label{fig:latency vs sharegpt}
    \vspace{-0.15in}\
    \end{subfigure}
    
    \caption{Impacts of datasets and models on overall serving systems.}
    \label{fig:latency-model-dataset}
\end{figure}


\mypar{Effectiveness of loading-optimized checkpoints}
We aimed to assess the effectiveness of loading-optimized checkpoints within a complete serverless workload, employing various model sizes and datasets to diversely test the checkpoint loaders.


\begin{figure}[t]
    \centering
     \begin{subfigure}{0.45\textwidth}
     \centering
    \includegraphics[width=\linewidth]{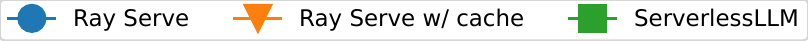}     
     \end{subfigure}
    \begin{subfigure}{0.49\linewidth}
    \includegraphics[width=\linewidth]{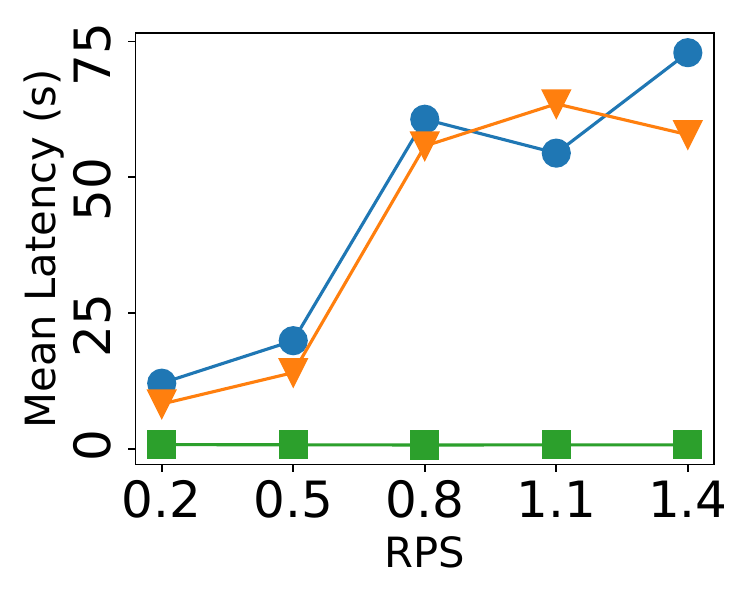}
    \vspace{-15pt}\
    \caption{GSM8k}
    \label{fig:latency-rps-gsm8k}
    \end{subfigure}
    \hfill
    \begin{subfigure}{0.48\linewidth}
    \includegraphics[width=\linewidth]{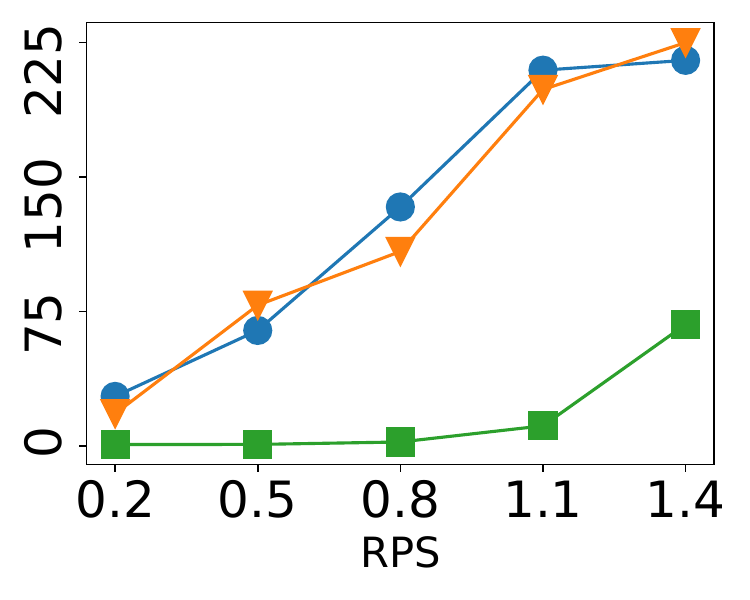}
    \vspace{-15pt}\
    \caption{ShareGPT }
    \label{fig:latency-rps-sharegpt}
    \end{subfigure}
    \caption{Impacts of RPS on overall serving systems.}
    \label{fig:latency-rps}
\end{figure}

In this experiment, as depicted in Figure~\ref{fig:latency-model-dataset}, Ray Serve and Ray Serve with Cache utilize Safetensors. Owing to the large sizes of the models, the SSD cache cannot accommodate all models, necessitating some to be downloaded from the storage server. With OPT-6.7B and GSM 8K, \sys starts models in an average of 0.8 seconds, whereas Ray Serve takes 12.1 seconds and Ray Serve with Cache 8.2 seconds, demonstrating an improvement of over 10X. Even with a faster network (\ie 100 Gbps), the average latency of Ray Serve could drop to 3.8 seconds, making it still 4.7 times slower than \sys. The significance of the model loader becomes more pronounced with larger models, as \sys can utilize parallel PCIe links when loading large models partitioned on multiple GPUs from pinned memory pool. For instance, with OPT-30B, \sys still initiates the model in 7.5 seconds, while Ray Serve's time escalates to 213 seconds and Ray Serve with Cache to 199.2 seconds, marking a 28X improvement.

This considerable difference in latency substantially affects the user experience in LLM services. Our observations indicate that \sys can fulfill 89\% of requests within a 300-second timeout with OPT-30B, whereas Ray Serve with Cache manages only 26\%.

With the ShareGPT dataset (Figure~\ref{fig:latency vs sharegpt}), which incurs a 3.7X longer inference time than GSM 8K, the challenge for model loaders becomes even more intense. For models like 6.7B and 13B, \sys achieves latencies of 0.8 and 1.6 seconds on average, respectively, compared to Ray Serve and Ray Serve with Cache, which soar to 182.2 and 162.4 seconds. When utilizing OPT-30B, \sys begins to confront GPU limitations (with all GPUs occupied and migration unable to free up more resources), leading to an increased latency of 89.9 seconds. However, this is still a significant improvement over Ray Serve with Cache, which reaches a latency of 261.8 seconds

\mypar{Effectiveness of live migration and loading scheduler}
In evaluating the effectiveness of LLM live migration and the loading scheduler, we created workloads with varying RPS levels. Scenarios with higher RPS highlight the importance of achieving load balancing and locality-aware scheduling since simply speeding up model loading is insufficient to address the resource contention common at large RPS levels.

From Figure~\ref{fig:latency-rps-gsm8k}, it is evident that \sys, equipped with GSM8K, consistently maintains low latency, approximately 1 second, even as RPS increases. In contrast, both Ray Serve and Ray Serve with Cache experience rising latency once the RPS exceeds 0.5, which can be attributed to GPU resource shortages. Their inability to migrate LLM inference for locality release or to achieve load balancing, unlike \sys, results in performance degradation.

With the more demanding ShareGPT workload, as shown in Figure~\ref{fig:latency-rps-sharegpt}, \sys maintains significant performance improvements — up to 212 times better — over Ray Serve and Ray Serve with Cache across RPS ranging from 0.2 to 1.1. However, at an RPS of 1.4, \sys's latency begins to rise, indicating that despite live migration and optimized server scheduling, the limited GPU resources eventually impact \sys's performance.

\begin{figure}[t]
    \centering
     \begin{subfigure}{0.45\textwidth}
     \centering
    \includegraphics[width=\linewidth]{figs//ray_evaluation/legend-serverlessllm.pdf}
     \end{subfigure}
     
    \begin{subfigure}{0.239\textwidth}
    \includegraphics[width=\linewidth]{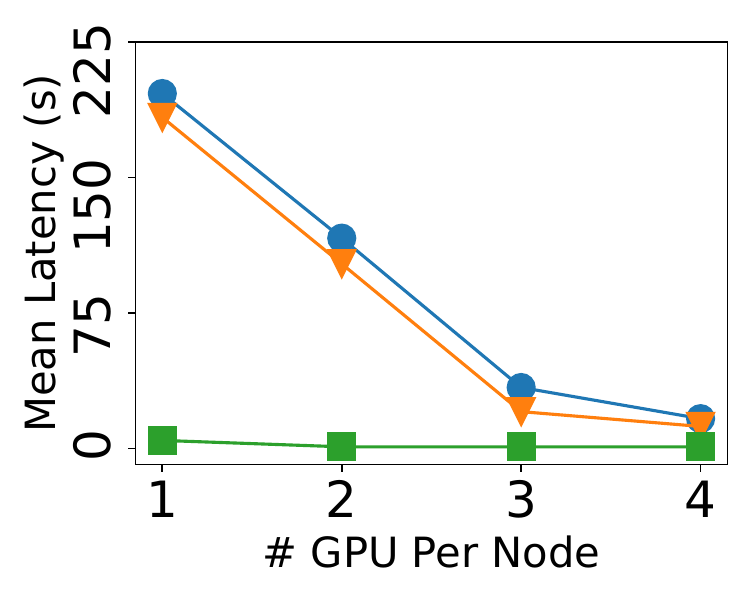}
    \caption{Impacts of \# GPUs per node}
    \label{fig:latency-gpu}
    \end{subfigure}
    \hfill
    \begin{subfigure}{0.232\textwidth}
    \includegraphics[width=\linewidth]{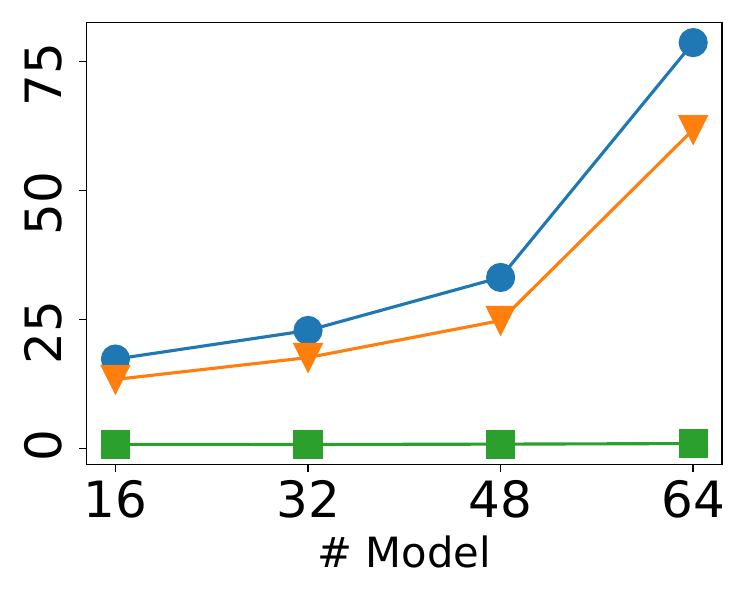}
    \caption{Impacts of \# models}
    \label{fig:latency-model}
    \end{subfigure}
    \caption{System scalability and resource efficiency.}
    \label{fig:latency-scaling}
\end{figure}

\mypar{Resource efficiency}
A major advantage of the low model startup latency in \sys is its contribution to resource savings when serving LLMs. We vary the number of GPUs available on each server to represent different levels of resource provisioning. As shown in Figure~\ref{fig:latency-gpu}, \sys scales well with elastic resources. With just one GPU per server, \sys already achieves a 4-second latency by efficient migrations and swaps. In contrast, Ray Serve with Cache requires at least four GPUs per server to attain a 12-second latency, which is still higher than \sys's performance with only one GPU per node. With larger clusters, the resource-saving efficiency of \sys is expected to become even more pronounced, as larger clusters offer more options for live migration and server scheduling. 

The resource efficiency of \sys is further evident when maintaining a fixed number of GPUs while increasing the number of LLMs in the cluster. In Figure~\ref{fig:latency-model}, with a limited number of models, Ray Serve with Cache can match \sys in latency performance. However, as the number of models grows, the performance gap widens, showcasing \sys's potential suitability for large-scale serverless platforms.

\mypar{KServe comparison} 
In our study, we assess KServe and \sys within a Kubernetes cluster. Given that our four-server cluster is unsuitable for a Kubernetes deployment, we instead utilize an eight-GPU server, simulating four nodes with two GPUs each. Since KServe performs slower than the other baselines considered in our evaluation, we only briefly mention KServe's results without delving into details. 

With KServe, the GPU nodes initially exhibited a first token latency of 128 seconds. This latency was primarily due to KServe taking 114 seconds to download an OPT-6.7B model checkpoint from the local S3 storage over a 1 Gbps network.
However, after applying the same enhancement as those for Ray Serve, we reduced the first token latency to 28 seconds.
Despite this improvement, KServe's best latency was significantly higher than those achieved by \sys. 
Notably, \sys was the only system able to reduce the latency to within one second.

\section{Related Work}

\mypar{Serverless inference systems}
Extensive research has focused on optimizing ML model serving in serverless architectures, targeting batching~\cite{234998, ali2020batch, 10.1145/3503222.3507709}, scheduling~\cite{yu2021gillis, romero2021infaas}, and resource efficiency~\cite{li2022tetris, gpulets-ml-inference}. Industry solutions like AWS SageMaker and Azure ML~\cite{azureml}, along with the open-source KServe~\cite{kserve}, demonstrate practical implementations. Despite these advancements, serverless inference systems still perform suboptimally with LLMs, as our paper demonstrates.

\mypar{Serverless cold-start optimizations}
Cold-start latency is a significant issue in serverless systems, addressed through various strategies including fast image pulling~\cite{273798}, lightweight isolation~\cite{216031, DBLP:conf/usenix/LiC0GBTZWHG22}, snapshot and restore~\cite{du2020catalyzer, 10.1145/3492321.3524270, 10.1145/3342195.3392698, 10.1145/3445814.3446714, 254432}, resource pre-provision~\cite{254430}, elastic resource provisioning~\cite{mai2020kungfu, wagenlander2020spotnik}, and fork~\cite{288627, 215935}. These approaches mainly focus on reducing startup times for containers or VMs without loading large external states. Recent research has explored optimizing cold-starts by facilitating faster model swaps between GPUs and host memory~\cite{DBLP:journals/corr/abs-2306-03622, eurosys23servinggpudirecthostaccess}, though scalability with LLMs is still challenging. In contrast, \sys effectively minimizes cold-start latency through LLM-specific innovations, such as optimized checkpoint formats and loading pipelines, live migration, and a cluster scheduler tailored to LLM inference characteristics.

\mypar{Exploiting locality in serverless systems}
Locality plays a crucial role in various optimization strategies for serverless systems. This includes leveraging host memory and local storage for data cache~\cite{222563, 10.1145/3472883.3486974, 10.14778/3407790.3407836}, optimizing the reading of shared logs~\cite{10.1145/3477132.3483541}, and enhancing communication efficiency in serverless Directed Acyclic Graphs (DAGs)~\cite{273835, 280890}. \sys, distinct from existing methods, introduces a high-performance checkpoint cache for GPUs, markedly improving checkpoint loading from multi-tier local storage to GPU memory. Recent studies~\cite{10.1145/3552326.3567496, 285113} have also recognized the need for leveraging locality in orchestrating serverless functions. Beyond these studies, \sys leverages LLM-specific characteristics in improving the locality-based server's selection and launching locality-driven inference.

\mypar{LLM serving systems}
Recent advancements in LLM serving have improved inference latency and throughput. Orca~\cite{280922} uses continuous batching for better GPU utilization during inference. AlpaServe~\cite{DBLP:conf/osdi/LiZZL00HCZGS23} shows that model parallelism can enhance throughput while meeting SLO constraints, though it has yet to be tested on generative models. vLLM~\cite{vllm} introduces PagedAttention for efficient KV cache management. SplitWise~\cite{patel2023splitwise} improves throughput by distributing prompt and token generation phases across different machines. Some approaches~\cite{10.5555/3571885.3571946, pmlr-v202-sheng23a} also use storage devices to offload parameters from GPUs to manage large LLM sizes. However, these systems often overlook model loading challenges, leading to increased first token latencies when multiple models share GPUs. \sys addresses this by focusing on minimizing loading latency to complement these throughput and latency optimizations.

\section{Conclusion}
This paper describes \sys, a low-latency serverless inference system purposefully designed for LLMs.
The design of \sys uncovers significant opportunities for system research, including designing new loading-optimized checkpoints, discovering the need to support live migration when conducting locality-driven LLM inference, and enabling a serverless cluster scheduler to be aware of the locality of checkpoints in a cluster when optimizing its model scheduling decision. We believe our work can be extended to ensure fairness of resources across the cluster and explore the possibility of smart checkpoint placement. We look forward to addressing these issues in the future. We consider \sys as the first step towards unlocking the potential of serverless computing for LLMs. We will continue to develop the open-source version of \sys. Given its versatility, we envision it as a platform to test new research ideas.

\section{Acknowledgments}

We sincerely thank our shepherd, Amar Phanishayee, and the OSDI reviewers for their insightful feedback, which helped improve the quality of this paper. We also extend our thanks to Boris Grot for his feedback. We acknowledge Zhaonan Zhang, Mingyu Dai, and Yusen Fei for their work in the early stages of this project. We are grateful for the GPU resources provided by the School of Informatics and the Edinburgh International Data Facility. This work is supported by UK EPSRC and gifts from Tencent.

\bibliographystyle{plain}
\bibliography{reference}

\end{document}